\documentclass[10pt,twocolumn,letterpaper]{article}

\usepackage{cvpr}              
\usepackage{multirow}
\usepackage{multicol}
\usepackage{xcolor}
\usepackage{pifont}
\usepackage{colortbl}
\usepackage{comment}
\usepackage{tcolorbox}
\usepackage{verbatim}
\usepackage[accsupp]{axessibility}

\definecolor{cvprblue}{rgb}{0.21,0.49,0.74}
\usepackage[pagebackref,breaklinks,colorlinks,allcolors=cvprblue]{hyperref}

\def\paperID{7349} 
\def\confName{CVPR}
\def\confYear{2025}

\newcommand{\methodname}{MaskCLIP}
\newcommand{\taskname}{OpenSDI}
\newcommand{\datasetname}{OpenSDID}

\def\zhiwu{\textcolor{black}}
\def\ZH{\textcolor{black}}
\def\yabin{\textcolor{black}}

\title{OpenSDI: Spotting Diffusion-Generated Images in the Open World}

\author{Yabin Wang
\\
Xi'an Jiaotong University\\
{\tt\small iamwangyabin@stu.xjtu.edu.cn}
\and
Zhiwu Huang\thanks{Corresponding author.}\\
University of Southampton\\
{\tt\small Zhiwu.Huang@soton.ac.uk}
\and
Xiaopeng Hong\\
Harbin Institute of Technology\\
{\tt\small hongxiaopeng@ieee.org}
}

\begin{document}
\maketitle
\begin{abstract}

This paper identifies OpenSDI, a challenge for spotting diffusion-generated images in open-world settings. In response to this challenge, we define a new benchmark, the OpenSDI dataset (OpenSDID), which stands out from existing datasets due to its diverse use of large vision-language models that simulate open-world diffusion-based manipulations. Another outstanding feature of OpenSDID is its inclusion of both detection and localization tasks for images manipulated globally and locally by diffusion models. To address the OpenSDI challenge, we propose a Synergizing Pretrained Models (SPM) scheme to build up a mixture of foundation models. This approach exploits a collaboration mechanism with multiple pretrained foundation models to enhance generalization in the OpenSDI context, moving beyond traditional training by synergizing multiple pretrained models through prompting and attending strategies. Building on this scheme, we introduce MaskCLIP, an SPM-based model that aligns Contrastive Language-Image Pre-Training (CLIP) with Masked Autoencoder (MAE). Extensive evaluations on OpenSDID show that MaskCLIP significantly outperforms current state-of-the-art methods for the OpenSDI challenge, achieving remarkable relative improvements of 14.23\% in IoU (14.11\% in F1) and 2.05\% in accuracy (2.38\% in F1) compared to the second-best model in \ZH{localization and detection} tasks, respectively. \ZH{Our dataset and code are available at \url{https://github.com/iamwangyabin/OpenSDI}}.


\end{abstract}    
\section{Introduction}
\label{sec:intro}

\begin{figure}
    \centering    \includegraphics[width=1\linewidth]{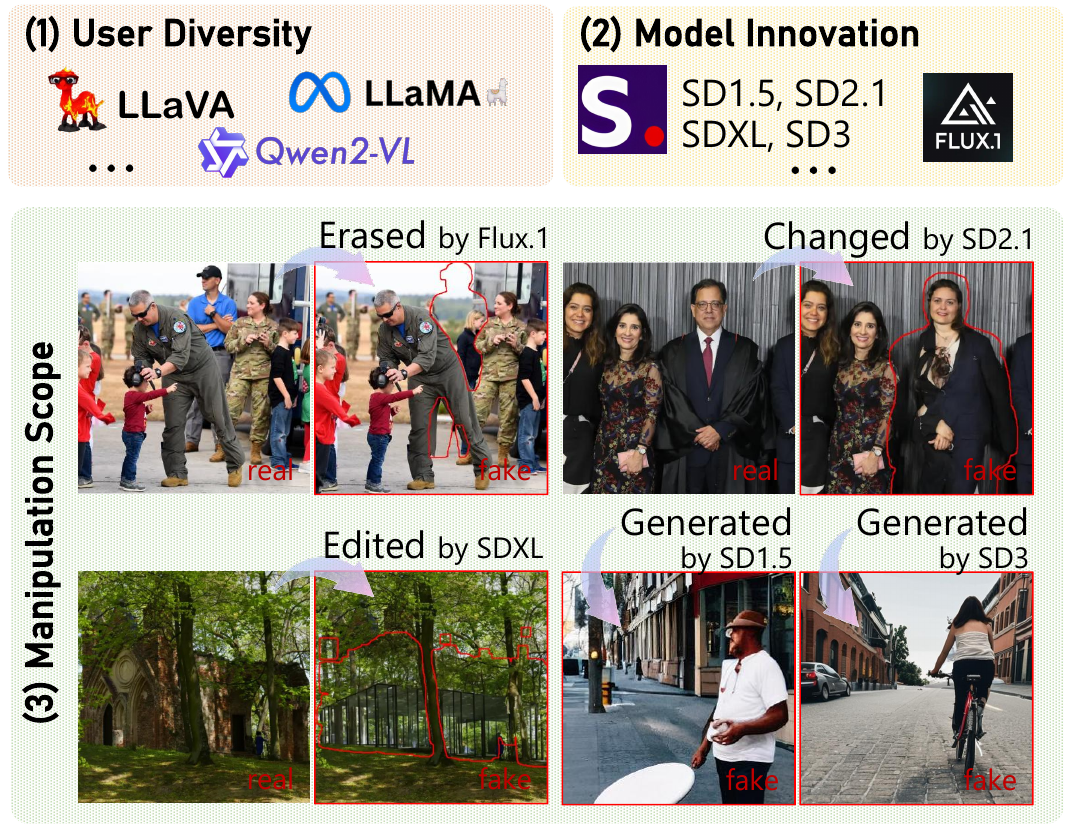}
    \caption{Three open-world settings of the OpenSDI challenge: (1) user diversity by simulating a range of user preferences, (2) model innovation through the use of multiple advanced diffusion models, and (3) a full manipulation scope that enables both global and local generation.}
    \label{fig:challenge}
    \vspace{-0.2cm}
\end{figure}



The rise of advanced text-to-image (T2I) generation models, such as Stable Diffusion~\cite{rombach2022high,esser2024scaling, podell2023sdxl} and other diffusion-based frameworks~\cite{flux2024, saharia2022photorealistic, xue2024raphael}, has profoundly transformed digital content creation, enabling users to produce remarkably lifelike images from simple text prompts. This capability, while expanding creative possibilities, has blurred the boundary between real and AI-generated content, making it increasingly difficult to ensure content authenticity. Addressing this challenge requires detectors that can handle three essential open-world settings (Figure~\ref{fig:challenge}): (1) the wide range of user preferences (\emph{user diversity}), covering diverse styles, subjects, and creative intentions; (2) the rapid evolution of diffusion models (\emph{model innovation}), introducing significant variability in generated images; and (3) the extensive T2I generation (\emph{manipulation scope}), from a broad spectrum of global image synthesis to precise local changes, object insertions, and background manipulations~\cite{zhang2023adding, ruiz2023dreambooth, zhao2024uni}. We identify this challenge as \emph{Open-world Spotting of Diffusion Images (\taskname)}.

In response to the OpenSDI challenge, we introduce \datasetname, a large-scale dataset specifically designed to address the scarcity of suitable public datasets for benchmarking \taskname ~(\ZH{Table~\ref{tab:dataset_comparison}}). Our approach incorporates the three essential dimensions of OpenSDI to create a comprehensive and realistic dataset.
To reflect \emph{user diversity}, we use multiple large vision-language models (VLMs)~\cite{wang2024qwen2, liu2023llava, llama3.2vision2024, li2023scaling} to generate a broad range of natural language prompts, simulating how users iteratively refine their creative ideas through diverse, open-world textual descriptions. These prompts are designed to mirror real-world user behavior and preferences, capturing the variety seen in different creative intentions and styles. 
To address \emph{model innovation}, we incorporate state-of-the-art (SOTA) T2I diffusion models~\cite{rombach2022high, podell2023sdxl, esser2024scaling, flux2024}, each with distinct architectures and parameter configurations. This variety in diffusion models allows our dataset to encompass the unique visual characteristics and nuances associated with advanced diffusion models, enhancing the dataset’s relevance for evaluating detection and localization across a spectrum of T2I diffusion models.
Finally, for \emph{manipulation scope}, we leverage the semantic understanding of VLMs in combination with Segment Anything Model (SAM)~\cite{kirillov2023segment} and Florence 2~\cite{xiao2023florence}  to translate textual instructions into precise region masks. These masks enable a wide range of manipulations, from local region edits to global image synthesis, simulating the types of modifications that make detection challenging in open-world settings. By combining global and localized manipulations, \datasetname ~aims at embodying the complexity required for benchmarking the \taskname ~challenge.


To develop effective approaches for addressing OpenSDI, we face two main barriers: (1) unifying detection and localization frameworks, and (2) generalizing to open-world settings. The unification is challenging due to the high heterogeneity of detection and location. Detection relies on image-level semantics, whereas localization requires pixel-level precision. To tackle these issues, we suggest an approach of building up a mixture of pretrained foundation models. While employing multiple expert models holds significant potential, achieving synergy among them remains a complex challenge. To address this, we introduce a novel framework, Synergizing Pretrained Models (SPM). It is designed to optimize the collaboration between pretrained foundation models for enhanced performance and maintaining generalization capability across both detection and localization tasks in open-world settings. In particular, this approach leverages the strengths of multiple pretrained models through prompt tuning and attention mechanisms, creating a mixture of pretrained foundation models that enhances generalization across diverse real-world scenarios. Building on this SPM approach, we introduce MaskCLIP, a mixture of pretrained models that aligns the capabilities of Contrastive Language-Image Pre-Training (CLIP) with a Masked Autoencoder (MAE) model, outperforming SOTA techniques with considerable relative improvements in both detecting and localizing diffusion-generated image content on the suggested OpenSDI dataset.

In summary, our main contributions are as follows: 
\begin{itemize}
\item We define the OpenSDI challenge, and introduce a large-scale dataset (OpenSDID) designed to address the lack of suitable public datasets for OpenSDI.
\item We introduce an SPM framework and its derived MaskCLIP to effectively address the OpenSDI challenge.
\item Through extensive experiments, MaskCLIP demonstrates markedly superior performance compared to SOTA methods for OpenSDI.
\end{itemize}

\section{Related Works}
\label{sec:rw}

\begin{table}
\centering
\resizebox{1\linewidth}{!}{

\begin{tabular}{l|rr|c|c|c}
\hline
Dataset & \# Real & \# Fake & User  & Model & Full \\ \hline
DiffusionDB~\cite{wang2023diffusiondb} & -- & 14M  & \textcolor{green}{\ding{51}}  & \textcolor{red}{\ding{55}}  & \textcolor{red}{\ding{55}}  \\ 
GenImage~\cite{zhu2024genimage} & 1.3M & 1.4M  & \textcolor{red}{\ding{55}}  & \textcolor{green}{\ding{51}}  & \textcolor{red}{\ding{55}}  \\ 
AutoSplice~\cite{jia2023autosplice} & 2.3K & 3.6K  & \textcolor{red}{\ding{55}}  & \textcolor{red}{\ding{55}}  & \textcolor{green}{\ding{51}}  \\ 
CocoGlide~\cite{guillaro2023trufor} & -- & 512  & \textcolor{red}{\ding{55}}  & \textcolor{red}{\ding{55}} & \textcolor{red}{\ding{55}}  \\ 
HiFi-IFDL~\cite{guo2023hierarchical} & -- & 1M*  & \textcolor{green}{\ding{51}}  & \textcolor{red}{\ding{55}}  & \textcolor{green}{\ding{51}}  \\ 
GIM~\cite{chen2024gim}  & 1.1M & 1.1M & \textcolor{red}{\ding{55}}  & \textcolor{green}{\ding{51}}  & \textcolor{red}{\ding{55}}  \\ 
TGIF~\cite{mareen2024tgif} & 3.1K & 75K & \textcolor{red}{\ding{55}}  & \textcolor{green}{\ding{51}}  & \textcolor{red}{\ding{55}}   \\ \hline
\textbf{OpenSDID} & 300K & 450K & \textcolor{green}{\ding{51}}  & \textcolor{green}{\ding{51}}  & \textcolor{green}{\ding{51}}  \\  \hline
\end{tabular}
}
\caption{Overview of existing diffusion image datasets and the proposed OpenSDID. User: Simulates diverse preferences. Model: Leverages multiple advanced diffusion models published after 2023. Full: Includes both global and local generation. Note: The asterisk (*) indicates HiFi-IFDL is built from multiple prior datasets. A more comprehensive version is in the suppl. material.}
\label{tab:dataset_comparison}
\end{table}

\noindent\textbf{Diffusion Image Benchmark Datasets.}
Existing datasets fall short in addressing the \taskname{} challenge due to their significant limitations.
Many focus on traditional manipulation techniques such as splicing and copy-move operations. 
Widely used datasets like Columbia~\cite{hsu2006detecting}, NIST MFC~\cite{guan2019mfc}, and CASIA~\cite{dong2013casia} concentrate on these conventional methods. HiFi-IFDL~\cite{guo2023hierarchical} created a million-scale dataset that combines various generative models and datasets. 
However, the manipulation types in HiFi-IFDL are mainly centered around traditional methods and facial editing, and do not involve recent T2I diffusion generators. 

To the best of our knowledge, only a few datasets incorporate text-guided editing using diffusion models such as DiffusionDB~\cite{wang2023diffusiondb}, GenImage~\cite{zhu2024genimage}, CocoGlide~\cite{guillaro2023trufor}, AutoSplice~\cite{jia2023autosplice}, TGIF~\cite{mareen2024tgif} and GIM~\cite{chen2024gim}. DiffusionDB uses a single diffusion model, while
CocoGlide and AutoSplice are relatively small in scale and restrict their scope to specific models (Glide~\cite{nichol2022glide} and DALL-E2~\cite{ramesh2022hierarchical}). 
While TGIF and GIM offer a broader range of diffusion models, they lack diversity in user preferences as their editing instructions come from a single source. Furthermore, these datasets primarily focus on partial edits rather than comprehensive image generation, limiting their scope. 
As shown in Table~\ref{tab:dataset_comparison}, none of these datasets fully addresses all three essential aspects of the \taskname{} challenge. 


\noindent\textbf{Detection and Localization Approaches.}
Early detection methods~\cite{wang2020cnn, chai2020makes} simply use CNN-based architecture, such as ResNet~\cite{he2016deep} and Xception~\cite{chollet2017xception}, as binary classifiers for distinguishing between real and synthetic images.
Several works~\cite{frank2020leveraging, tan2024frequencyaware, tan2023rethinking, liu2020global, tan2023learning} exploit distinctive frequency patterns that differentiate synthetic from real images to improve model generalization performance. 
However, these methods often struggle with generalization across different types of generators, as they tend to overfit to specific artifacts present in their training data. 
Recent methods~\cite{ojha2023towards, wang2022s, liu2022detecting, wang2023dire, khan2024clipping} employ  large pretrained models, particularly CLIP~\cite{radford2021learning}, to construct binary classifiers for deepfake detection.
While they evince promising detection performance, they are limited to binary predictions and lack the ability to precisely localize tampered regions. 

Another family of approaches (e.g.,~\cite{wu2021iid, wu2019mantra, hu2020span}) provides both detection results and accurate localization of manipulated regions.
Early methods like~\cite{chen2021image, liu2022pscc, guillaro2023trufor, kwon2022learning} design complex feature extraction and fusion modules across spatial, frequency, and noise domains to enhance the precision of forged image localization. 
Recently, vision transformer based methods~\cite{han2024hdf, ma2024imlvit, wang2022objectformer} clearly improved the accuracy and efficiency of detection and localization. 
While these approaches primarily rely on ImageNet pretrained models, newer innovations such as~\cite{10.1007/978-3-031-73247-8_2, lai2023detect} are pushing boundaries by leveraging large-scale vision-language models like CLIP for more robust image forgery detection and localization.

However, most existing methods focus on traditional image manipulations, overlooking those generated globally and locally by advanced T2I diffusion models, leaving a substantial gap in addressing \taskname{}. Furthermore, they have limited generalization when faced with new types of AI-generated images, underscoring a clear limitation in adapting to rapidly evolving image generation technologies.

\section{Dataset Creation}

We introduce \datasetname, a large-scale dataset specifically curated for the OpenSDI challenge. Our dataset design addresses the three core requirements essential for open-world spotting of AI-generated content: user diversity, model innovation, and manipulation scope.

\noindent\textbf{User diversity.} To simulate the variety of user intentions and styles in real-world settings, we employ an automated pipeline driven by multiple state-of-the-art VLMs, including LLaMA3 Vision~\cite{llama3.2vision2024}, LLaVA~\cite{liu2023llava}, InternVL2~\cite{chen2024far} and Qwen2 VL~\cite{wang2024qwen2}. These VLMs simulate human-like editing behavior, generating diverse text prompts that mirror the wide range of real-world image manipulation intentions. Based on an authentic set of images from Megalith-10M~\cite{BoerBohan2024Megalith10m}, which is a resource of wholesome, unedited, copyright-free licensed images\footnote{\yabin{The images are licensed under: United States Government Work, Public Domain Dedication (CC0), No known copyright restrictions (Flickr commons), or Public Domain Mark.}}, our pipeline utilizes VLMs to produce varied manipulation instructions. These prompts emulate common user-driven edits, such as adding or removing objects, altering backgrounds, or changing key image elements. This approach ensures that \datasetname{} accurately captures the breadth of \yabin{user-like} diversity in T2I generation, producing a dataset with rich and varied samples that reflect real-world creative diversity.

\noindent\textbf{Model Innovation.} As diffusion models evolve rapidly, it is crucial to account for the diverse visual styles and architectural innovations of different T2I diffusion models. To achieve this, we leverage a suite of state-of-the-art generative models—including multiple versions of Stable Diffusion (SD1.5~\cite{rombach2022high}, SD2.1~\cite{rombach2022high}, SDXL~\cite{podell2023sdxl}, SD3~\cite{esser2024scaling}) and Flux.1~\cite{flux2024}. 
These models span various architectures, parameter configurations, and generation characteristics, providing our dataset with a wide range of model-specific image outputs. By using different models to execute the manipulations specified in the VLM-generated prompts, we ensure that \datasetname~ covers a comprehensive spectrum of T2I outputs, reflecting the progression and variety within diffusion-based image synthesis. Our dataset includes distinct manipulation styles and visual attributes unique to each model version, supporting more rigorous testing of detection systems across evolving T2I diffusion models.

\noindent\textbf{Manipulation Scope.} As illustrated in Figure~\ref{fig:challenge}, we leverages T2I diffusion models for both global image synthesis and precise local edits. To simulate this full range of manipulations, our pipeline integrates SAM~\cite{kirillov2023segment} and Florence 2~\cite{xiao2023florence} to translate textual prompts into detailed region masks, enabling accurate localized edits. For each image, SAM and Florence generate segmentation masks that delineate specific areas for modifications.
\yabin{During the editing phase, we use the AutoPipelineForInpainting\footnote{\url{https://huggingface.co/docs/diffusers/using-diffusers/inpaint}}, a standard practice in image edition, 
to ensure that only the masked regions are modified.} 
To further enhance dataset diversity, we produce multiple variations of each manipulation by using different random seeds. 
To enhance dataset quality and ensure that generated images align precisely with intended instructions, we employ the CLIP model for optimal image selection. 
We further diversify our dataset by incorporating traditional editing techniques, such as copy-paste and splicing operations~\cite{mareen2024tgif}, applied at random intervals. 
Additionally, we include fully synthetic images to ensure comprehensive coverage of diffusion image types.
This fusion of global and local generation ensures that \datasetname~  captures the complexity of real-world T2I forgeries, covering a comprehensive manipulation spectrum.

\begin{figure}
    \centering\includegraphics[width=1\linewidth]{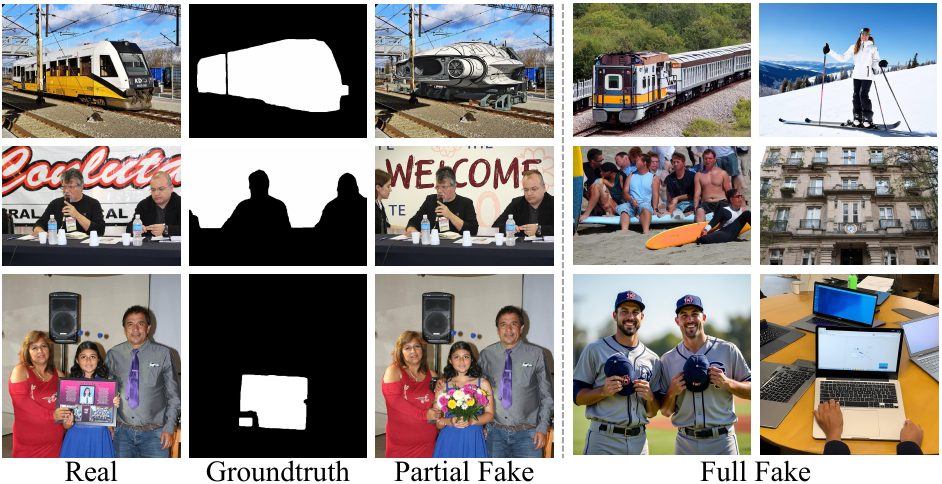}
    \caption{OpenSDID examples of authentic and manipulated images with their corresponding ground-truth masks, and entire generated images. 
    }
    \label{fig:some-samples}
    \vspace{-0.3cm}

\end{figure}

The dataset statistics are presented in Table~\ref{tab:dataset_stats}, with some samples provided in Figure~\ref{fig:some-samples}. 
OpenSDID comprises 300,000 images, evenly distributed between real and fake samples, with 200,000 images (100,000 real \zhiwu{from Megalith-10M} and 100,000 fake \zhiwu{generated by SD1.5}) used for training, and 100,000 images (20,000 each from SD1.5, SD2.1, SDXL, SD3, and Flux.1) reserved for testing, ensuring a balanced representation across multiple generative models.
\yabin{This design follows common practices in AI-generated image detection~\cite{guillaro2023trufor,wang2020cnn}, evaluating generalization to unseen generators.}
Additional analysis and details about this dataset, including considerations of ethical and bias issues, are provided in the suppl. material.

\begin{table}
\centering
\begin{tabular}{l|rr|rr|r}
\hline
\multirow{2}{*}{Model} & \multicolumn{2}{c|}{Training Set} & \multicolumn{2}{c|}{Test Set} & Total \\
& Real & Fake & Real & Fake & Images \\
\hline
\textbf{SD1.5} & 100K & 100K & 10K & 10K & 220K \\
\textbf{SD2.1} & - & - & 10K & 10K & 20K \\
\textbf{SDXL} & - & - & 10K & 10K & 20K \\
\textbf{SD3} & - & - & 10K & 10K & 20K \\
\textbf{Flux.1} & - & - & 10K & 10K & 20K \\
\hline
\textbf{Total} & 100K & 100K & 50K & 50K & 300K \\
\hline
\end{tabular}
\caption{Dataset Statistics. }
\label{tab:dataset_stats}
\vspace{-0.3cm}
\end{table}

\section{Proposed Approach}

\noindent\textbf{SPM Framework.} To tackle the OpenSDI challenge, we focus on two primary issues: (1) unifying detection and localization frameworks, and (2) generalizing to open-world environments. Detection and localization are inherently heterogeneous—the former relies on image-level semantics to classify an image, while the latter demands pixel-level accuracy to identify specific manipulated regions. To address them, we propose a scheme of Synergizing Pretrained Models (SPM) to optimize collaboration among pretrained models, each excelling in learning image- or pixel-level semantics while showing strong generalization capabilities.


The suggested SPM scheme achieves this synergy through two key mechanisms: \emph{prompting} and \emph{attending}. The prompting strategy introduces an additional parametrization to the network input, enabling the model to encode new knowledge without significantly altering the pretrained network components. This approach typically involves minimal fine-tuning or even freezing the original foundation model, ensuring that the pretrained knowledge is largely preserved. The attending strategy, on the other hand, facilitates collaboration between multiple pretrained models via a cross-attention mechanism, aligning feature representations from different models to improve performance. We demonstrate the effectiveness of the SPM scheme with MaskCLIP, which is based on a mixture of CLIP for image-level semantic understanding and MAE for fine-grained spatial precision. 




\begin{figure*}[h]
    \centering
    \includegraphics[width=0.95\linewidth]{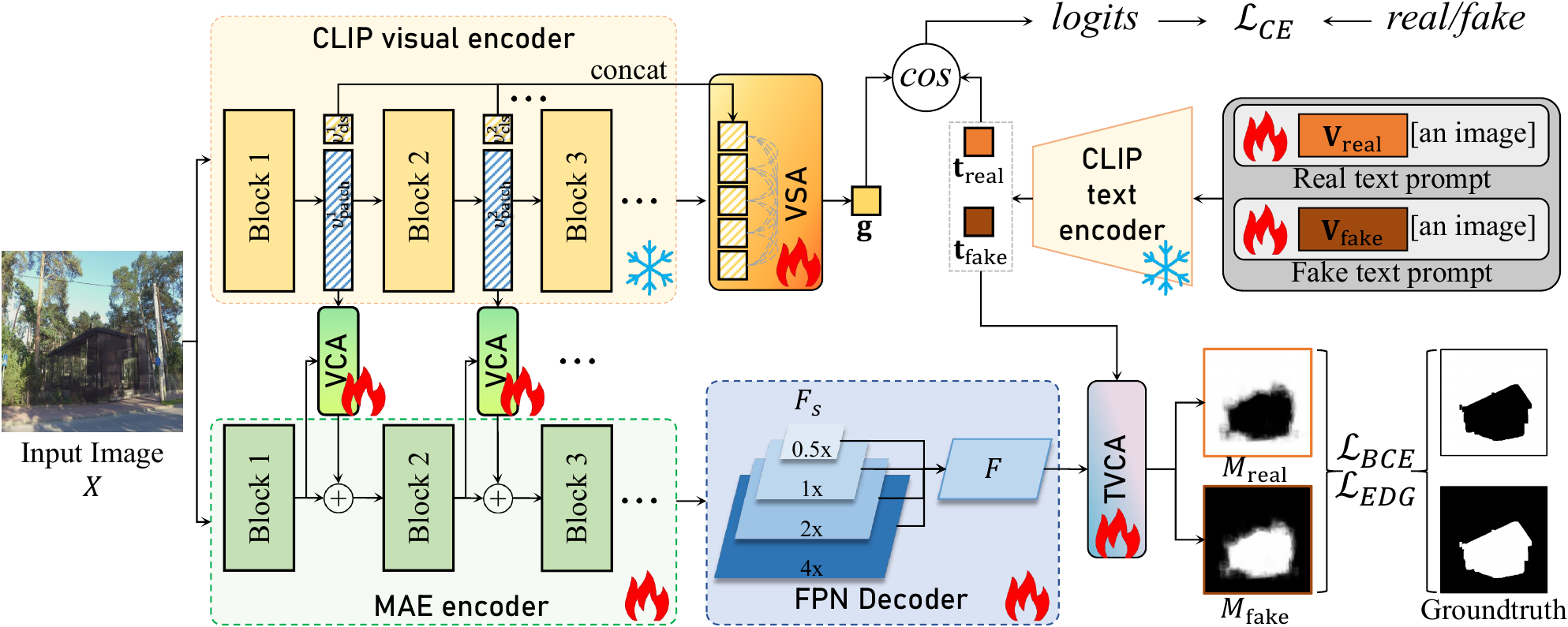}
    \caption{MaskCLIP overview. MaskCLIP inherits core components from pretrained CLIP and MAE: the CLIP vision and text encoders, the MAE encoder, and uses an \zhiwu{FPN-style decoder for precise pixel-level predictions.} To synergize CLIP and MAE, MaskCLIP introduces one prompting block (top right) and three attention blocks: VCA (visual cross-attention), TVCA (textual-visual cross-attention), and VSA (visual self-attention). Frozen components are marked with snowflakes, while tunable components (marked with flames) are optimized using \zhiwu{a balanced objective with $\mathcal{L}_{\text{CE}}$ (cross-entropy), $\mathcal{L}_{\text{BCE}}$ (binary cross-entropy), and $\mathcal{L}_{\text{EDG}}$ (edge-weighted loss)}.}
    \label{fig:framework}
\end{figure*}

\noindent\textbf{Preliminaries.} The CLIP~\cite{radford2021learning} model is a dual-encoder architecture consisting of a text encoder $E_t$ and an vision encoder $E_v$. 
The pretrained CLIP has demonstrated impressive zero-shot classification performances by computing cosine similarity between the visual features extracted from images and text features generated from class descriptions.

Given an input image $X \in \mathbb{R}^{H \times W \times 3}$ and text description $T$, the vision encoder splits the input image $X$ into $N$ image patches $\{x_i\}_{i=1}^N$.
Each patch $x_i$ is flattened and projected to obtain visual tokens $v_i \in \mathbb{R}^D$. 
This set of input visual tokens can be defined as $V_{patch} = \{v_i\}_{i=1}^N$. 
A learnable [CLS] token $v_{cls} \in \mathbb{R}^D$ is prepended to the sequence of visual tokens to form the complete input sequence, to facilitate a global image representation.
Thus the input sequence for layer $l$ is formed by concatenating the CLS token and visual tokens as $V^l = [v_{cls}^l; V_{patch}^l] \in \mathbb{R}^{(N+1) \times D}$.

Through the self-attention of the transformer, the $v_{cls}$ token iteratively aggregates information from all visual tokens, resulting in a global representation of the entire image by the final layer.
The [CLS] token $v_{cls}^L$ from the final layer $L$ is linearly projected to get the global image representation $\mathbf{g}$.
Similarly, the text encoder embeds the text description $T$ through the transformer to produce text embeddings $\mathbf{t}$. 
The image and text feature $\mathbf{g}$ and $\mathbf{t}$ are then normalized and used to compute the cosine similarity between image-text pairs.

MAEs~\cite{he2022masked} are self-supervised vision transformers that learn rich visual representations through a reconstruction objective of masked image regions.
Unlike CLIP, the MAE encoder $E_m$ processes the input image $X$ without a [CLS] token, generating only visual patch tokens $V_m = \{v_{m,i}\}_{i=1}^N$ where $v_{m,i} \in \mathbb{R}^{D_m}$. 
The decoder attempts to reconstruct the original image patches corresponding to the masked tokens, encouraging the model to learn meaningful features from the spatial context available in the visible tokens.

\noindent\textbf{MaskCLIP Overview.} As shown in Figure~\ref{fig:framework}, MaskCLIP integrates the CLIP visual encoder, the CLIP text encoder, and the MAE encoder. \zhiwu{Following the MAE study~\cite{he2022masked}, we replace the vanilla MAE decoder with a Feature Pyramid Network (FPN)~\cite{lin2017feature} based decoder, as done in ~\cite{ma2024imlvit}, for more precise pixel-level predictions.} Based on the SPM scheme, we align the CLIP and MAE components through \emph{prompting} and \emph{attending} to address the OpenSDI challenge. Rather than full fine-tuning, we employ a prompt-tuning strategy~\cite{khattak2023maple, wang2022s, jia2022visual} with the textual prompting block (Figure~\ref{fig:framework}), learning a set of continuous prompt vectors for real/fake concepts while preserving CLIP’s original parameters to retain its extensive visual-linguistic knowledge. To effectively integrate the prompted CLIP with MAE, the \emph{attending} process is critical. For this purpose, we introduce two types of attending blocks (Figure~\ref{fig:framework}): (1) visual cross-attention (VCA), which aligns the CLIP visual encoding features with the MAE encoding features, and (2) textual-visual cross-attention (TVCA), which coordinates the CLIP textual encoding features with the enhanced MAE decoding features. Additionally, we add a visual self-attention (VSA) block to concatenate global features from each CLIP encoding layer for more robust feature extraction. \zhiwu{During training, all the prompting and attending blocks (VCA, TVCA, VSA) are jointly optimized.}

\noindent\textbf{Prompting.} 
MaskCLIP learns a pair of continuous prompt vectors $\mathbf{V}_c \in \mathbb{R}^{M \times D}$ for $c \in \{\text{real}, \text{fake}\}$, where $M$ denotes the prompt length and $D$ represents the embedding dimension. These learned prompts function analogously to hand-crafted text prompts like ``\texttt{a photo of a [real/fake]}'', but employ learnable continuous vectors to capture the semantic context more effectively. 
These learnable prompts $\mathbf{V}_c$ are then processed by the text encoder to generate class-specific embeddings $\mathbf{t}_{\text{real}}$ and $\mathbf{t}_{\text{fake}} \in \mathbb{R}^D$, representing the real and fake class concepts, respectively.

\noindent\textbf{Attending.}  
To align CLIP with MAE, \methodname{} employs the VCA block that combines CLIP's semantic understanding with MAE's spatial reconstruction capabilities. VCA is deployed across multiple network layers to facilitate hierarchical feature fusion between CLIP and MAE.
At each fusion layer $l$, VCA processes spatial features $V^l_m$ from the MAE encoder alongside the corresponding CLIP visual token features $V_{patch}^l$. 
To align these features, we first apply bilinear interpolation to resize $V_{patch}^l$ to the spatial size of MAE features.
Subsequently, a $1 \times 1$ convolution is employed to project the feature dimension $D$ to $D_m$.
The aligned CLIP features serve as query vectors $\mathbf{Q}$, while the MAE features $V^l_m$ are projected to form keys and values $\mathbf{K}, \mathbf{V} \in \mathbb{R}^{N \times D_m}$, respectively. 
We use the multi-head attention mechanism within VCA is computed as:
\begin{equation}
\ZH{\mathbf{G}}^l = \text{Softmax}\left( \frac{\mathbf{Q} \mathbf{K}^\top}{\sqrt{d}} \right)\mathbf{V},
\vspace{-0.2cm}
\end{equation}
where $d$ is the feature dimension for the attention head. To enhance the fusion features, a residual connection is introduced, and the input features for layer $l+1$ of the MAE is  $V^{l+1}_m = V^l_m + \ZH{\mathbf{G}}^l$.

In order to further align the CLIP textual features with the decoding features $F$, we suggest incorporating a TVCA block to integrate textual information into the decoding process. 
Given a textual embedding ${\mathbf{t}_{\text{real}}, \mathbf{t}_{\text{fake}}}$, the decoder performs cross-attention between $\mathbf{t}$ and \zhiwu{the decoder's output feature $F$}. First, $F$ is linearly projected to match the dimension of $\mathbf{t}$. 
The cross-attention operation uses $\mathbf{t}$ as queries and $F$ as keys and values to get the attention feature map $F'$.
Finally we use a $1 \times 1$ convolutional layer to generate the segmentation logits $M_{\text{fake}}$ and $M_{\text{real}} \in \mathbb{R}^{H \times W}$.

In addition, the conventional approaches~\cite{ojha2023towards, wang2022s} of utilizing only the final layer's [CLS] token $v_{cls}^L$ from the CLIP vision encoder fails to fully leverage the rich hierarchical features learned throughout the network. 
This limitation is particularly critical for AI-generated image detection, where both low-level artifacts and high-level semantic inconsistencies play crucial roles in discrimination. Thus, rather than relying solely on the last layer output [CLS] token $v_{cls}^L$ of the CLIP vision encoder, we introduce the VSA block to attend [CLS]s from multiple the CLIP encoder layers.
VSA leverages a multi-head self-attention mechanism that collects [CLS] token embeddings $v_{cls}^l \in \mathbb{R}^D$ from multiple layers $l \in L$, forming a sequence $V_{cls} = \{v_{cls}^l\}_{l \in L}$.
The output features are subsequently aggregated via global average pooling and transformed through a learned linear projection to derive the final image representation $\mathbf{g}$.

\noindent\textbf{\ZH{Essential Add-ons}.}
\zhiwu{Inspired by prior works~\cite{ma2024imlvit, guo2023hierarchical, he2022masked}, we employ an FPN-based decoder \cite{lin2017feature, he2017mask}} that processes the MAE encoder's output features $V^L_m$ at multiple scales (0.5×, 2×, 4×), enabling the capture of both fine and coarse details.
At each resolution, a $3 \times 3$ convolutional layer is applied to extract information. These multi-scale features $F_s$ are then resized to the original spatial shape $(H, W)$, and are concatenated along the channel dimension to form the feature map $F$.

To train MaskCLIP for localization, we adopt a combination of binary cross-entropy loss $\mathcal{L}_{\text{BCE}}$ and edge-weighted loss $\mathcal{L}_{\text{EDG}}$ \yabin{applied to $M_{\text{fake}}$}, following previous works~\cite{guillaro2023trufor, ma2024imlvit}. 
The total loss function incorporates the detection loss (i.e., cross-entropy loss term, $\mathcal{L}_{\text{CE}}$) to optimize the prompts, VSA, VCA, FPN Decoder, TVCA and the MAE encoder jointly.
The weights for all loss terms are set to $1$.

During inference, the binary detection is performed by the computing of cosine similarities between the image representation $\mathbf{g}$ and class-specific text embeddings ${\mathbf{t}_{\text{real}}, \mathbf{t}_{\text{fake}}}$.
In parallel, the TVCA block generates the forgery localization map $M_{\text{fake}}$ to predict potentially manipulated regions.

\section{Experiments}


\begin{table*}
\centering
\resizebox{1\linewidth}{!}{
\begin{tabular}{>{\columncolor{gray!10}}l|>{\columncolor{blue!10}}c>{\columncolor{blue!10}}c|>{\columncolor{red!10}}c>{\columncolor{red!10}}c|>{\columncolor{red!10}}c>{\columncolor{red!10}}c|>{\columncolor{red!10}}c>{\columncolor{red!10}}c|>{\columncolor{red!10}}c>{\columncolor{red!10}}c|>{\columncolor{gray!10}}c>{\columncolor{gray!10}}c} \hline
\multirow{2}{*}{} & \multicolumn{2}{c|}{\cellcolor{blue!10}SD1.5} & \multicolumn{2}{c|}{\cellcolor{red!10}SD2.1} & \multicolumn{2}{c|}{\cellcolor{red!10}SDXL} & \multicolumn{2}{c|}{\cellcolor{red!10}SD3} & \multicolumn{2}{c|}{\cellcolor{red!10}Flux.1} &
\multicolumn{2}{c}{\cellcolor{gray!10}AVG} \\
Method & IoU & F1 & IoU & F1 & IoU & F1 &  IoU & F1 &  IoU & F1 & IoU & F1 \\
\hline
MVSS-Net~\cite{chen2021image} & 0.5785 & 0.6533 & 0.4490 & 0.5176 & 0.1467 & 0.1851 & 0.2692 & 0.3271 & 0.0479 & 0.0636 & 0.2983 & 0.3493 \\
CAT-Net~\cite{kwon2022learning} & 0.6636 & \underline{0.7480} & 0.5458 & \underline{0.6232} & 0.2550 & 0.3074 & \underline{0.3555} & \underline{0.4207} & 0.0497 & 0.0658 & \underline{0.3739} & \underline{0.4330} \\
PSCC-Net~\cite{liu2022pscc} & 0.5470 & 0.6422 & 0.3667 & 0.4479 & 0.1973 & 0.2605 & 0.2926 & 0.3728 & 0.0816 & 0.1156 & 0.2970 & 0.3678 \\
ObjectFormer~\cite{wang2022objectformer} & 0.5119 & 0.6568 & 0.4739 & 0.4144 & 0.0741 & 0.0984 & 0.0941 & 0.1258 & 0.0529 & 0.0731 & 0.2414 & 0.2737 \\
TruFor~\cite{guillaro2023trufor} & 0.6342 & 0.7100 & \underline{0.5467} & 0.6188 & \underline{0.2655} & \underline{0.3185} & 0.3229 & 0.3852 & 0.0760 & 0.0970 & 0.3691 & 0.4259 \\
DeCLIP~\cite{smeu2024declip} & 0.3718 & 0.4344 & 0.3569 & 0.4187 & 0.1459 & 0.1822 & 0.2734 & 0.3344 & \underline{0.1121} & \underline{0.1429} & 0.2520 & 0.3025 \\
IML-ViT~\cite{ma2024imlvit} & \underline{0.6651} & 0.7362 & 0.4479 & 0.5063 & 0.2149 & 0.2597 & 0.2363 & 0.2835 & 0.0611 & 0.0791 & 0.3251 & 0.3730 \\
\hline
\methodname & \textbf{0.6712} & \textbf{0.7563} & \textbf{0.5550} & \textbf{0.6289} & \textbf{0.3098} & \textbf{0.3700} & \textbf{0.4375} & \textbf{0.5121} & \textbf{0.1622} & \textbf{0.2034} & \textbf{0.4271} & \textbf{0.4941} \\
\hline
\end{tabular}
}
\vspace{-0.2cm}
\caption{Pixel-level (localization) performance on \datasetname{}. Blue Block: in-domain test, Red Block: cross-domain test, \textbf{Bold}: best results, \underline{Underline}: second best results. More results are in the supplementary material.}
\label{table:pixel}
\vspace{-0.1cm}
\end{table*}

\begin{table*}
\centering
\resizebox{1\linewidth}{!}{
\begin{tabular}{>{\columncolor{gray!10}}l|>{\columncolor{blue!10}}c>{\columncolor{blue!10}}c|>{\columncolor{red!10}}c>{\columncolor{red!10}}c|>{\columncolor{red!10}}c>{\columncolor{red!10}}c|>{\columncolor{red!10}}c>{\columncolor{red!10}}c|>{\columncolor{red!10}}c>{\columncolor{red!10}}c|>{\columncolor{gray!10}}c>{\columncolor{gray!10}}c} \hline
\multirow{2}{*}{\cellcolor{gray!10}} & \multicolumn{2}{c|}{\cellcolor{blue!10}SD1.5} & \multicolumn{2}{c|}{\cellcolor{red!10}SD2.1} & \multicolumn{2}{c|}{\cellcolor{red!10}SDXL} & \multicolumn{2}{c|}{\cellcolor{red!10}SD3} & \multicolumn{2}{c|}{\cellcolor{red!10}Flux.1} & \multicolumn{2}{c}{\cellcolor{gray!10}AVG} \\
Method& F1 & Acc & F1 & Acc & F1 & Acc &  F1 & Acc &  F1 & Acc & F1 & Acc \\
\hline
CNNDet~\cite{wang2020cnn} & \underline{0.8460} & \underline{0.8504} & 0.7156 & 0.7594 & 0.5970 & 0.6872 & 0.5627 & 0.6708 & 0.3572 & 0.5757 & 0.6157 & 0.7087 \\
GramNet~\cite{liu2020global} & 0.8051 & 0.8035 & 0.7401 & 0.7666 & 0.6528 & 0.7076 & 0.6435 & 0.7029 & 0.5200 & 0.6337 & 0.6723 & 0.7229 \\
FreqNet~\cite{tan2024frequencyaware} & 0.7588 & 0.7770 & 0.6097 & 0.6837 & 0.5315 & 0.6402 & 0.5350 & 0.6437 & 0.3847 & 0.5708 & 0.5639 & 0.6631 \\
NPR~\cite{tan2023rethinking} & 0.7941 & 0.7928 & \underline{0.8167} & {0.8184} & \underline{0.7212} & 0.7428 & \textbf{0.7343} & \underline{0.7547} & \textbf{0.6762} & \textbf{0.7136} & \underline{0.7485} & \underline{0.7645} \\
UniFD~\cite{ojha2023towards} & 0.7745 & 0.7760 & 0.8062 & \underline{0.8192} & 0.7074 & \underline{0.7483} & 0.7109 & 0.7517 & \underline{0.6110} & \underline{0.6906} & 0.7220 & 0.7572 \\
RINE~\cite{10.1007/978-3-031-73220-1_23} & \textbf{0.9108} & \textbf{0.9098} & \textbf{0.8747} & \textbf{0.8812} & \textbf{0.7343} & \textbf{0.7876} & \underline{0.7205} & \textbf{0.7678} & 0.5586 & 0.6702 & \textbf{0.7598} & \textbf{0.8033} \\
\midrule    
MVSS-Net~\cite{chen2021image} & 0.9347 & 0.9365 & 0.7927 & 0.8233 & 0.5985 & 0.7042 & 0.6280 & 0.7213 & 0.2759 & 0.5678 & 0.6460 & 0.7506 \\
CAT-Net~\cite{kwon2022learning} & \textbf{0.9615} & \underline{0.9615} & 0.7932 & 0.8246 & 0.6476 & \underline{0.7334} & 0.6526 & \underline{0.7361} & 0.2266 & 0.5526 & 0.6563 & 0.7616 \\
PSCC-Net~\cite{liu2022pscc} & \underline{0.9607} & {0.9614} & 0.7685 & 0.8094 & 0.5570 & 0.6881 & 0.5978 & 0.7089 & \underline{0.5177} & \underline{0.6704} & 0.6803 & \underline{0.7676} \\
ObjectFormer~\cite{wang2022objectformer} & 0.7172 & 0.7522 & 0.6679 & 0.7255 & 0.4919 & 0.6292 & 0.4832 & 0.6254 & 0.3792 & 0.5805 & 0.5479 & 0.6626 \\
TruFor~\cite{guillaro2023trufor} & 0.9012 & \textbf{0.9773} & 0.3593 & 0.5562 & 0.5804 & 0.6641 & 0.5973 & 0.6751 & 0.4912 & 0.6162 & 0.5859 & 0.6978 \\
DeCLIP~\cite{smeu2024declip} & 0.8068 & 0.7831 & \underline{0.8402} & \underline{0.8277} & \underline{0.7069} & 0.7055 & \underline{0.6993} & 0.6840 & \underline{0.5177} & 0.6561 & \underline{0.7142} & 0.7313 \\
IML-ViT~\cite{ma2024imlvit} & 0.9447 & 0.7573 & 0.6970 & 0.6119 & 0.4098 & 0.4995 & 0.4469 & 0.5125 & 0.1820 & 0.4362 & 0.5361 & 0.5635 \\
\hline
\methodname & 0.9264 & 0.9272 & \textbf{0.8871} & \textbf{0.8945} & \textbf{0.7802} & \textbf{0.8122} & \textbf{0.7307} & \textbf{0.7801} & \textbf{0.5649} & \textbf{0.6850} & \textbf{0.7779} & \textbf{0.8198} \\
\hline
\end{tabular}
}
\vspace{-0.2cm}

\caption{Image-level (detection) performance on \datasetname{}. Blue Block: in-domain test, Red Block: cross-domain test, Top Section: image-level detection methods, Bottom Section: pixel-level detection and localization methods. \textbf{Bold}: best results, \underline{Underline}: second best results. More results are in the supplementary material.
}
\label{table:image}
\vspace{-0.3cm}
\end{table*}

\begin{figure*}
    \centering
    \includegraphics[width=0.9\linewidth]{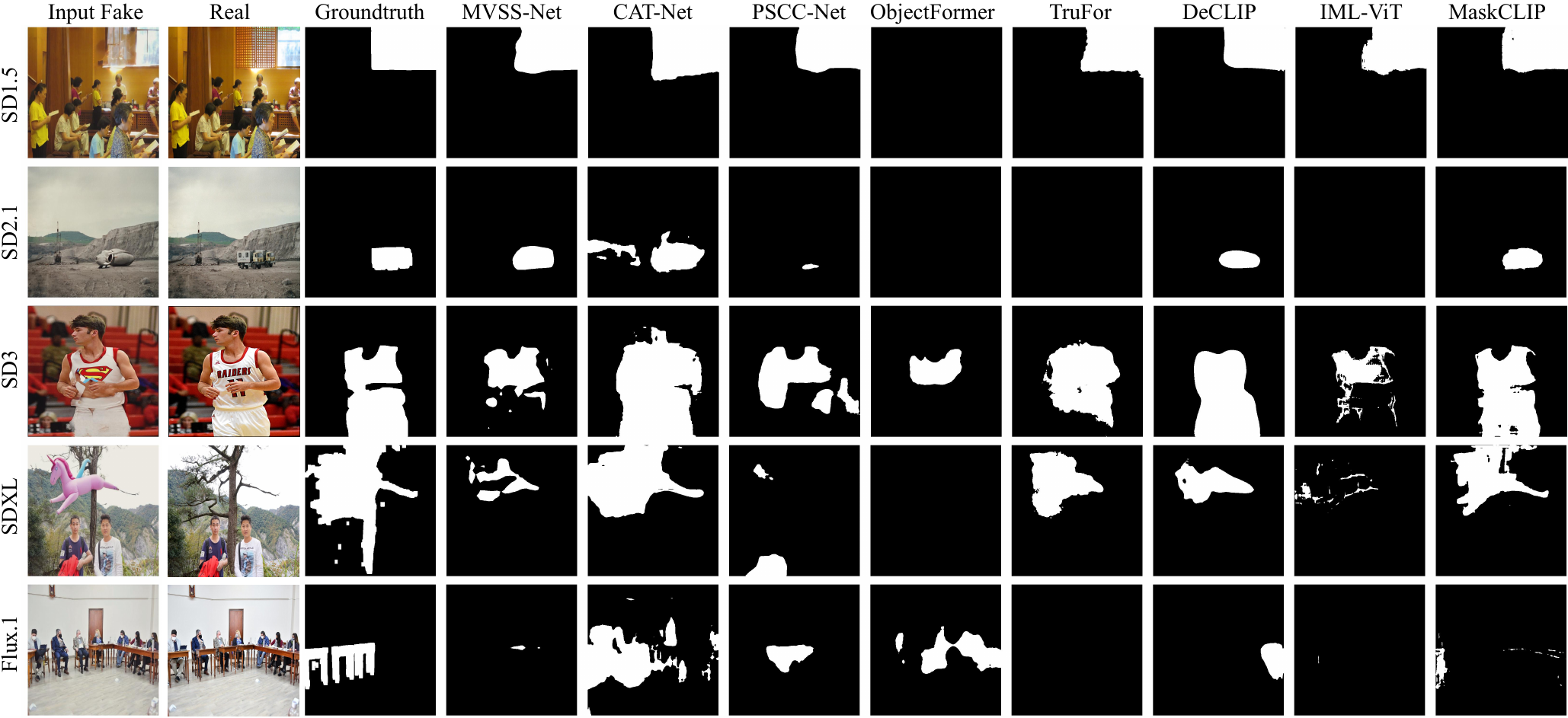}
    \caption{Qualitative results on OpenSDID. }
    \label{fig:qualitative}
\end{figure*}

\noindent\textbf{Implementation Details.}
The training of \methodname{} is conducted with batch size $64$ and learning rate $1e-4$ for $20$ epoch using the Adam optimizer. 
For data augmentation, we adopt a set of standard settings~\cite{ma2024imdl, guillaro2023trufor,ma2024imlvit, liu2022pscc}, including Gaussian blurring, JPEG compression, random scaling, horizontal and vertical flips, and random cropping to $512\times 512$. 
However, for the CLIP encoder, images are resized to $224\times 224$ pixels to maintain compatibility with its pretrained configuration.
In the main experiments, we choose the OpenAI ViT-L/14\footnote{We also explore CLIP variants
and alternative MAE models. To ensure reliability, we conduct experiments with varying hyperparameters and multiple runs. \zhiwu{More details are provided in the suppl. material.}} in line with the previous works~\cite{smeu2024declip, ojha2023towards, 10.1007/978-3-031-73220-1_23}. 
For the MAE encoder, we use the ImageNet pretrained weights ViT-B/32 as in previous works~\cite{ma2024imlvit}.
We set the prompt length $M$ to be $10$ as previous works~\cite{wang2022s, ren2024prompt}.
All training is conducted on $4$ NVIDIA H100 GPUs.


\noindent\textbf{State-of-the-art Methods.}
These methods include classical image manipulation localization methods, including CAT-Net~\cite{kwon2022learning}, MVSS-Net~\cite{chen2021image}, PSCC-Net~\cite{liu2022pscc} and TruFor~\cite{guillaro2023trufor}.
For transformer-based methods, we also include ObjectFormer~\cite{wang2022objectformer}, IML-ViT~\cite{ma2024imlvit} and DeCLIP~\cite{smeu2024declip}.
In addition to these pixel-level based methods, we also include SOTA image-level based (detection) methods, include CNNDet~\cite{wang2020cnn}, UniFD~\cite{ojha2023towards}, NPR~\cite{tan2023rethinking}, GramNet~\cite{liu2020global}, FreqNet~\cite{tan2024frequencyaware}, and RINE~\cite{10.1007/978-3-031-73220-1_23}.
All the methods were implemented using the IMDL-BenCo framework~\cite{ma2024imdl} and refer to their official codes if possible. 
We trained these methods on the \datasetname{}'s training data (Table \ref{tab:dataset_stats}), carefully tuning their hyperparameters to ensure optimal performance that matches their reported benchmarks.
A brief description of these methods is provided in the supplementary.


\noindent\textbf{Evaluation Metrics.} Following standard practices in~\cite{guillaro2023trufor, ma2024imdl}, we evaluate methods using both pixel-level and image-level metrics on the \datasetname{}'s test data (Table \ref{tab:dataset_stats}). For pixel-level evaluation, we employ the F1-score (F1) and Intersection over Union (IoU). At the image level, we assess performance using F1 and accuracy (Acc).


\noindent\textbf{Localization Results.}
In Table~\ref{table:pixel}, we show the pixel-level localization performance in terms of F1 and IoU metrics. 
For the in-domain localization, specifically on the SD1.5 subset, \methodname{} achieves an IoU of $0.6712$ and an F1 score of $0.7563$, surpassing all other methods. Notably, our method outperforms IML-ViT, the second best, by $\textbf{0.92\%}$ in IoU and $\textbf{2.73\%}$ in F1 score.
In terms of the cross-domain generalization, \methodname{} shows superior performance across various unseen generators. 
Our method outperforms previous best methods by significant margins: SD2.1 ($\textbf{1.52\%}$ IoU, $\textbf{0.91\%}$ F1), SDXL ($\textbf{16.69\%}$ IoU, $\textbf{16.17\%}$ F1), SD3 ($\textbf{23.07\%}$ IoU, $\textbf{21.73\%}$ F1), and most notably on Flux.1 ($\textbf{44.69\%}$ IoU, $\textbf{42.34\%}$ F1).
The average performance across all datasets further emphasizes the superiority of \methodname{}. It achieves an average IoU of 0.4271 and an average F1 score of 0.4941, surpassing the second best by $\textbf{14.23\%}$ in IoU and $\textbf{14.11\%}$ in F1. 
The markedly superior in-domain and cross-domain performance suggests that \methodname{} does not overfit to specific content or manipulation patterns but learns generalizable features crucial for detecting and localizing forgeries across diverse open-world scenarios.

\noindent\textbf{Detection Results.}
\yabin{Table~\ref{table:image} displays image-level detection performance (F1-score and Accuracy) across various generative models, separating methods focused solely on image-level detection from those addressing both image and pixel-level tasks. }
The results illustrate that \methodname{} achieves the best average performance among all evaluated methods, demonstrating both strong in-domain detection and robust cross-domain generalization capabilities.
For the in-domain detection \methodname{} shows competitive performance with the others.
However, in terms of the cross-domain generalization, \methodname{} exhibits clearly better performance than all the other methods. 
On the SD2.1 and SDXL subsets, MaskCLIP even outperforms the SOTA image-level deepfake detectors, RINE, by $\textbf{1.24\%}$ and $\textbf{4.59\%}$, respectively.
For the Flux.1 test data, while image-level detector NPR shows better performance compared to our multi-task detectors, \methodname{} achieves the best results among all localization methods.
For the average result, \methodname{} significantly outperforms all other methods by $\textbf{2.38\%}$ on F1 and by $\textbf{2.05\%}$ on Accuracy.

\noindent\textbf{Qualitative Comparisons.}
We show some output localization masks of \methodname{} and other SOTA approaches in Figure~\ref{fig:qualitative}. 
As shown across different generators, our method produces masks with well-defined object boundaries and shapes. 
For instance, the ground truth comparison demonstrates that our approach successfully captures vehicle shapes (SD2.1), clothing textures (SD3), and object shape (SDXL). In contrast, competing methods often exhibit limitations such as over-segmentation of objects or missed detections, highlighting the robustness of our framework.
More results are available in the suppl. materials.

\begin{figure}
    \centering\includegraphics[width=0.9\linewidth]{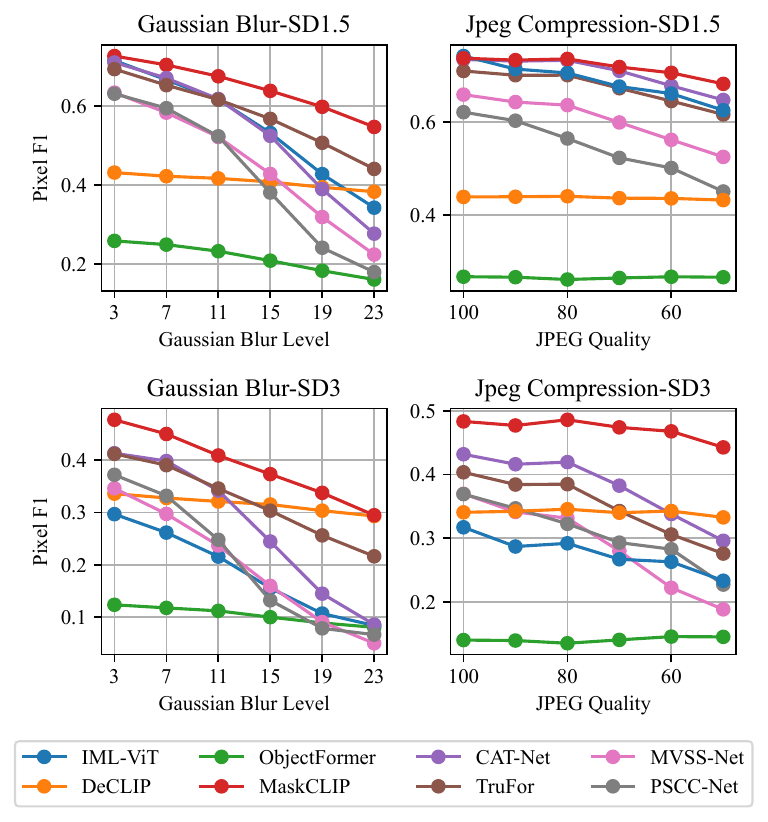}
    \vspace{-0.3cm}
    \caption{Robustness evaluation of different SOTA methods under image degradation. It compares performance across varying levels of Gaussian Blur (left) and JPEG Compression (right) on both in-domain (SD1.5) and cross-domain (SD3) test sets. \zhiwu{ Results on the rest test data are provided in the suppl. material.}}
    \label{fig:robustcompare}
    \vspace{-0.3cm}

\end{figure}

\noindent\textbf{Robustness Analysis.}
We conducted extensive experiments to evaluate detection robustness under different image degradation scenarios. 
Following standard practices~\cite{ma2024imdl, guillaro2023trufor}, we employed two common degradation methods: Gaussian blur and JPEG compression, applying them at various intensity levels. 
The results are presented in Figure~\ref{fig:robustcompare}, and we report the pixel-level F1 scores.
For in-domain evaluation (SD1.5), our MaskCLIP (shown in red line) demonstrates superior robustness across both degradation types. 
The proposed MaskCLIP exhibits strong robustness in cross-domain scenarios (SD3), though with a slightly reduced margin compared to in-domain results. 
This is particularly evident in the JPEG compression test, where our method maintains steady performance even at lower-quality settings. 
These comprehensive evaluations demonstrate our method's superior robustness in both in-domain and cross-domain scenarios, particularly under challenging image quality conditions.

\begin{table}
\centering
\resizebox{0.9\linewidth}{!}{
\begin{tabular}{l|c|c}
\hline
Model Variation & Pixel F1 & Image F1 \\
\hline
Baseline & 0.4290 & 0.7613 \\
+ Prompt Tuning & 0.4485 & 0.8576 \\
+ VSA & 0.4259 & 0.9083 \\
+ VCA & 0.6532 & 0.9343 \\
+ TVCA (Full \methodname) & 0.7563 & 0.9264 \\
\hline
\end{tabular}
}
\caption{Ablation study of the main components in \methodname{}. 
}
\label{tab:ablation}
\vspace{-0.3cm}
\end{table}

\noindent\textbf{Ablation Study.} To thoroughly evaluate the effectiveness of each component in our proposed method, we conduct comprehensive ablation experiments. Starting with a baseline architecture that combines UniFD's~\cite{ojha2023towards} classifier with DeCLIP's~\cite{smeu2024declip} decoder, we progressively incorporate our key components. 
Table~\ref{tab:ablation} presents the quantitative results on the \datasetname{} SD1.5 test set.
The baseline achieves a pixel-level F1 of 0.4290 and an image-level F1 of 0.7613. Prompt tuning improves image F1 to 0.8576, showing effective adaptation of CLIP representations. 
VSA further boosts image-level F1 to 0.9083. 
The VCA effectively combines CLIP's semantic and MAE's spatial features, which significantly enhances pixel-level F1. 
Finally, TVCA completes our \methodname{} model, achieving the highest pixel F1 of 0.7563 while maintaining strong image F1 at 0.9264. These results demonstrate that each component, particularly VCA and TVCA, contributes substantially to the model's overall performance.
\yabin{In the suppl. materials, we replace MAE with \ZH{SAM~\cite{kirillov2023segment}}, achieving a remarkable 0.7873 Pixel F1 on SD1.5, which further demonstrates the scalability of the proposed SPM scheme.}

\section{Conclusions and Outlook}

This paper addresses the OpenSDI challenge of detecting and localizing diffusion-generated images in open-world settings by presenting the OpenSDID benchmark dataset. OpenSDID stands out from existing datasets due to its use of diverse vision-language models and advanced diffusion models to replicate real-world local and global manipulations. To address this challenge, we proposed the approach of Synergizing Pretrained Models (SPM) and introduced MaskCLIP, an SPM-based model that synergizes CLIP and MAE.
\ZH{This enhances generalization at both image and pixel levels, achieving state-of-the-art detection and localization on OpenSDID.}
We envision OpenSDID as an open-ended dataset, capable of incorporating images generated by increasingly advanced generative AI models and serving as a standard benchmark for this evolving domain. 


\section{Acknowledgment}
This research was supported by the ECS Fund (504079140) at the University of Southampton, National Natural Science Foundation of China (62376070, 62076195), and the Fundamental Research Funds for the Central Universities (AUGA-5710011522). Yabin Wang was supported by the China Scholarship Council (CSC)  Grant No. 202306280130.
This work was primarily conducted at the University of Southampton.
The authors acknowledge the use of the IRIDIS X High Performance Computing Facility and the Southampton-Wolfson AI Research Machine (SWARM) GPU cluster, funded by the Wolfson Foundation, together with the associated support services at the University of Southampton in the completion of this work.

{
    \small
    \bibliographystyle{ieeenat_fullname}
    \bibliography{main}
}

\end{document}



\clearpage
\setcounter{page}{1}
\maketitlesupplementary

In this supplementary material, we provide additional details and results to complement the main paper. Section~\ref{sec:supp_creation} offers a step-by-step explanation of the OpenSDID creation pipeline, detailing the processes and tools used. 
Section~\ref{sec:dataset_details} includes a comprehensive analysis of the dataset's diversity and quality, along with a thorough discussion of the ethical and bias considerations. 
Section~\ref{sec:related_works_discussion} provides a more detailed comparison of the proposed \methodname{} with related works, highlighting its unique contributions and advantages.
To further validate the performance and robustness of \methodname{}, we present additional experimental results in Section~\ref{sec:more_results} to validate the performance and robustness of \methodname{}, including cross-dataset evaluations on traditional image forgery datasets and more recent AI-generated image benchmarks, robustness analysis under image degradation, and multiple runs to ensure stability. 
Additionally, we provide visualizations of the results and datasets in Sections~\ref{sec:sample1} and~\ref{sec:sample2}, offering intuitive insights into the data and the method's performance.
\ZH{We are committed to making our research reproducible and accessible to the broader community. Therefore, we have made both the OpenSDID dataset  and the complete codebase publicly available at \url{https://github.com/iamwangyabin/OpenSDI}.} 


\begin{figure*}
    \centering\includegraphics[width=1\linewidth]{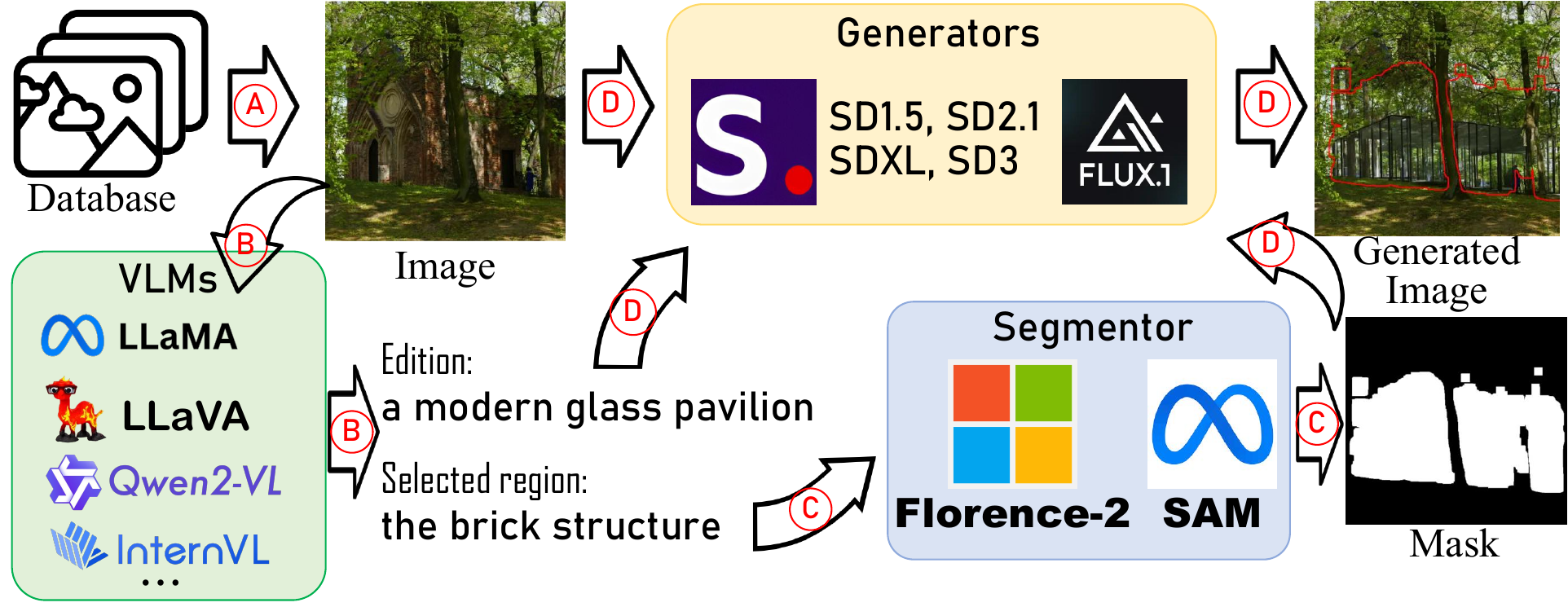}
    \caption{\zhiwu{An OpenSDID pipeline for local modification on real image content: (A) Sampling real images from the Megalith-10M dataset, (B) Generating textual instructions for editing using Vision Language Models (VLMs), (C) Creating visual masks for modification through segmentation models, and (D) Producing AI-generated images with image generators based on the instructions and masks. For global image content generation, OpenSDID merely uses (B) and (D) without using real images to produce masks. }}
    \label{fig:autopipeline}
\end{figure*}

\section{\datasetname{} Creation Pipeline}
\label{sec:supp_creation}

To create our \datasetname{}, we designed a comprehensive framework to automatically generate a large and diverse set of edited images using advanced text-to-image (T2I) diffusion generative models. 
As illustrated in Figure~\ref{fig:autopipeline}, the process consists of four key steps: (A) load real images from a database, (B) generate textual instructions for editing, (C) create visual masks for editing, and (D) produce AI-generated images with instructions and masks. 
For global image image generation, \datasetname{} merely uses (B) and (D) without using masks to generate images.

We begin by randomly selecting an authentic image from the Megalith-10M database~\cite{BoerBohan2024Megalith10m}. Next, we randomly choose a state-of-the-art large vision-language model, such as LLaMA3 Vision~\cite{llama3.2vision2024}, LLaVA~\cite{liu2023llava}, InternVL2~\cite{chen2024far}, or Qwen2 VL~\cite{wang2024qwen2}, to generate detailed edit suggestions. 
These models are selected for their ability to produce high-quality and contextually relevant edit instructions, ensuring that the subsequent image manipulations are both creative and realistic.
The prompts used to generate these edit suggestions are carefully crafted to guide the model in producing high-quality and relevant modifications. 
We emphasizes the importance of considering a wide range of potential modifications, including face transformations, hair modifications, body alterations, clothing and accessory changes, object replacements, background change, style modifications, and other adjustments. Below is an example of the  prompts that are likely to be used:

\begin{tcolorbox}[title=Prompt]
\begin{verbatim}
You are an image editor with decades 
of experience in digital manipulation 
and a reputation for innovation. 
Your expertise spans across various 
genres,including portrait retouching, 
architectural visualization, 
product photography, and surreal 
composites. Your task is to analyze 
image descriptions and propose 
compelling, realistic edits that 
could dramatically enhance or 
transform the image in unexpected 
yet believable ways.

When presented with an image, 
consider a comprehensive range of 
potential modifications. 
These modifications could be:
• Face transformations
• Hair modifications
• Body alterations
• Clothing and accessory changes
• Object replacements or additions
• Background transformations
• Architectural style modifications
• Vehicle transformations
• Food alterations and adjustments

Your suggestions should be both 
imaginative and feasible, 
taking into account the original 
image's context, composition, 
and lighting. Strive for a balance 
between creativity and photorealism, 
ensuring that your proposed edits 
could theoretically be executed 
by a skilled retoucher. 
\end{verbatim}
\end{tcolorbox}

The output text edit instructions specify the selected region in the given real images and the desired content to modify that area.
Thus, based on the edit instructions for the previous step, we utilize the Florence-2~\cite{xiao2023florence} and SAM~\cite{kirillov2023segment} models to obtain precise mask regions through open vocabulary detection and segmentation. The Florence-2 model identifies specific regions within the image using text input, while the SAM model refines these regions by generating more precise masks.
These masks are further processed by removing small disconnected components and expanded to ensure comprehensive coverage of the areas designated for inpainting.

One of the state-of-the-art T2I diffusion models (SD1.5~\cite{rombach2022high}, SD2.1~\cite{rombach2022high}, SDXL~\cite{podell2023sdxl}, SD3~\cite{esser2024scaling}) and Flux.1~\cite{flux2024}) is then employed to generate new images based on the masked regions and the corresponding prompts. The employed diffusion model takes three inputs: the original image, the generated mask, and the prompt describing the desired new content. To enhance the generation diversity, we randomly adjust key parameters, including the number of inference steps, guidance scale, and strength.

Finally, we use CLIP~\cite{radford2021learning} to compute similarity scores of the generated images and the given edit instructions. Only images that achieve high similarity scores are included in the final dataset, ensuring the maintenance of high-quality standards.

\section{OpenSDID Details and Analysis}
\label{sec:dataset_details}

\subsection{Comparison with Existing Datasets}

We give a more comprehensive comparison of the existing datasets in Table~\ref{tab:dataset_comparison}.
In terms of user-like diversity, existing datasets have often suffered from limited variability in user input due to their reliance on standardized generation pipelines. This strategy results in reduced diversity across generated images. While DiffusionDB made progress by collecting generated images from public Discord channels with varied generator parameters, it remained confined to Stable Diffusion 1.5 and globally generated images. 
Our dataset addresses these limitations by introducing substantial user-like diversity through the integration of multiple state-of-the-art Visual Language Models (VLMs), including LLaMA3 Vision and LLaVA. These VLMs simulate a broad spectrum of human-like editing behaviors, generating diverse text prompts that authentically reflect real-world manipulation intentions. 
Additionally, we randomize the generation hyperparameters during the diffusion process, including sampling steps, guidance scale, and seed values, to further enhance the diversity of our generated images.

Regarding Model Innovation, many existing datasets, such as DFFD~\cite{dang2020detection} and HiFi-Net~\cite{guo2023hierarchical}, primarily focus on images generated by GANs and early diffusion models, lacking representation of recent technological advances.
\datasetname{} addresses this limitation by incorporating multiple cutting-edge T2I diffusion models, including various versions of Stable Diffusion~\cite{rombach2022high,podell2023sdxl,esser2024scaling} and Flux.1~\cite{flux2024}.
This comprehensive inclusion provides a more robust platform for evaluating detection methods against the rapidly evolving landscape of image synthesis technologies.

With respect to Manipulation Scope, existing datasets often limit themselves to either global or local manipulations. For example, DiffusionDB~\cite{wang2023diffusiondb} and GenImage~\cite{zhu2024genimage} contain only globally synthesized images without local edits, whereas \datasetname{} uniquely combines both global and local manipulations. This comprehensive manipulation spectrum more accurately reflects real-world scenarios, where forgeries frequently involve complex combinations of global and localized alterations.

\begin{table*}[htbp]
\centering
\begin{tabular}{l|r|rr|r|c|c|c|c|c}
\hline
Dataset & Type & \# Real & \# Fake & Generator & Adv. T2I & Glo.  & Loc. & Users & \# Model \\ \hline
UADFV~\cite{yang2019exposing} & Face & 241 & 252 & GAN & \textcolor{red}{\ding{55}} & \textcolor{red}{\ding{55}} & \textcolor{green}{\ding{51}} & \textcolor{red}{\ding{55}} & 1 \\ 
DFFD~\cite{dang2020detection} & Face & 58K & 240K & GAN & \textcolor{red}{\ding{55}} & \textcolor{green}{\ding{51}} & \textcolor{green}{\ding{51}} & \textcolor{red}{\ding{55}} & 7 \\ 
FaceForensics++~\cite{rossler2018faceforensics} & Face & 1K & 4K & GAN & \textcolor{red}{\ding{55}} & \textcolor{red}{\ding{55}} & \textcolor{green}{\ding{51}} & \textcolor{red}{\ding{55}} & 1 \\ 
DFDC~\cite{dolhansky2020deepfake} & Face & 19K & 100K & GAN & \textcolor{red}{\ding{55}} & \textcolor{red}{\ding{55}} & \textcolor{green}{\ding{51}} & \textcolor{red}{\ding{55}} & 2 \\ 
DeeperForensics~\cite{jiang2020deeperforensics1} & Face & 50K & 10K & GAN & \textcolor{red}{\ding{55}} & \textcolor{red}{\ding{55}} & \textcolor{green}{\ding{51}} & \textcolor{red}{\ding{55}} & 1 \\ 
CNNSpot~\cite{wang2020cnn} & General & 362K & 362K & GAN & \textcolor{red}{\ding{55}} & \textcolor{green}{\ding{51}} & \textcolor{red}{\ding{55}} & \textcolor{red}{\ding{55}} & 13 \\ 
GenImage~\cite{zhu2024genimage} & General & 1.3M & 1.4M & GAN \& Diff. & \textcolor{green}{\ding{51}} & \textcolor{green}{\ding{51}} & \textcolor{red}{\ding{55}} & \textcolor{red}{\ding{55}} & 8 \\ 
DiffusionDB~\cite{wang2023diffusiondb} & General & - & 14M & Diff. & \textcolor{green}{\ding{51}} & \textcolor{green}{\ding{51}} & \textcolor{red}{\ding{55}} & \textcolor{green}{\ding{51}} & 1 \\ 
Columbia~\cite{ng2004data} & General & 933 & 912 & Trad. & \textcolor{red}{\ding{55}} & \textcolor{red}{\ding{55}} & \textcolor{green}{\ding{51}} & \textcolor{red}{\ding{55}} & - \\ 
CASIA~\cite{dong2013casia} & General & 7.2K & 5.1K & Trad. & \textcolor{red}{\ding{55}} & \textcolor{red}{\ding{55}} & \textcolor{green}{\ding{51}} & \textcolor{red}{\ding{55}} & - \\ 
IMD2020~\cite{novozamsky2020imd2020} & General & 35K & 35K & Trad. & \textcolor{red}{\ding{55}} & \textcolor{red}{\ding{55}} & \textcolor{green}{\ding{51}} & \textcolor{red}{\ding{55}} & - \\ 
NIST16~\cite{guan2019mfc} & General & - & 564 & Trad. & \textcolor{red}{\ding{55}} & \textcolor{red}{\ding{55}} & \textcolor{green}{\ding{51}} & \textcolor{red}{\ding{55}} & - \\ 
Coverage~\cite{wen2016coverage} & General & 100 & 100 & Trad. & \textcolor{red}{\ding{55}} & \textcolor{red}{\ding{55}} & \textcolor{green}{\ding{51}} & \textcolor{red}{\ding{55}} & - \\ 
AutoSplice~\cite{jia2023autosplice} & General & 2.3K & 3.6K & Diff. & \textcolor{green}{\ding{51}} & \textcolor{green}{\ding{51}} & \textcolor{green}{\ding{51}} & \textcolor{red}{\ding{55}} & 1 \\ 
CocoGlide~\cite{guillaro2023trufor} & General & -- & 512 &  Diff. & \textcolor{green}{\ding{51}} & \textcolor{red}{\ding{55}} & \textcolor{green}{\ding{51}} & \textcolor{red}{\ding{55}} & 1 \\ 
Dolos \cite{tantaru2024weakly} & Face & 20K & 105K & Diff. & \textcolor{green}{\ding{51}} & \textcolor{green}{\ding{51}} & \textcolor{green}{\ding{51}} & \textcolor{red}{\ding{55}} & 4 \\ 
HiFi-Net~\cite{guo2023hierarchical}* & General & -- & 1M & GAN \& Diff. & \textcolor{red}{\ding{55}} & \textcolor{green}{\ding{51}} & \textcolor{green}{\ding{51}} & \textcolor{red}{\ding{55}} & 10 \\ 
GIM~\cite{chen2024gim} & General & 300K & 1.1M & Diff. & \textcolor{green}{\ding{51}} & \textcolor{red}{\ding{55}} & \textcolor{green}{\ding{51}} & \textcolor{red}{\ding{55}} & 3 \\ 
TGIF~\cite{mareen2024tgif} & General & 3.1K & 75K & Diff. & \textcolor{green}{\ding{51}} & \textcolor{red}{\ding{55}} & \textcolor{green}{\ding{51}} & \textcolor{red}{\ding{55}} & 3 \\ \hline
\datasetname{} & General & 300K & 450K & Diff. & \textcolor{green}{\ding{51}} & \textcolor{green}{\ding{51}} & \textcolor{green}{\ding{51}} & \textcolor{green}{\ding{51}} & 5 \\ \hline
\end{tabular}
\caption{\zhiwu{Overview of more comprehensive image forgery datasets}. "Type" indicates the content category (Face or General). "\# Real \& \# Fake" indicates the number of real and fake images. "Generator" indicates synthesis method type (GAN, Diffusion (Diff.), or Traditional (Trad.)). "Adv. T2I" indicates whether advanced Text-to-Image models (e.g., Stable Diffusion) are used. "Glo." indicates the global image manipulations. "Loc." indicates the local image manipulations. "Users" indicates whether multiple users participate in dataset creation. "\# Model" indicates the number of distinct models used for generation. HiFi-Net's locally manipulated images are primarily sourced from previous traditional manipulation datasets and facial deepfake detection datasets.}
\label{tab:dataset_comparison}
\end{table*}

\subsection{Ethical and Bias Issues}

The proposed OpenSDID research has been approved by the University's Ethics and Research Governance Online team. Furthermore, we provide a more detailed analysis of potential ethical and bias issues in the dataset.

\begin{figure}
    \centering
    \includegraphics[width=1\linewidth]{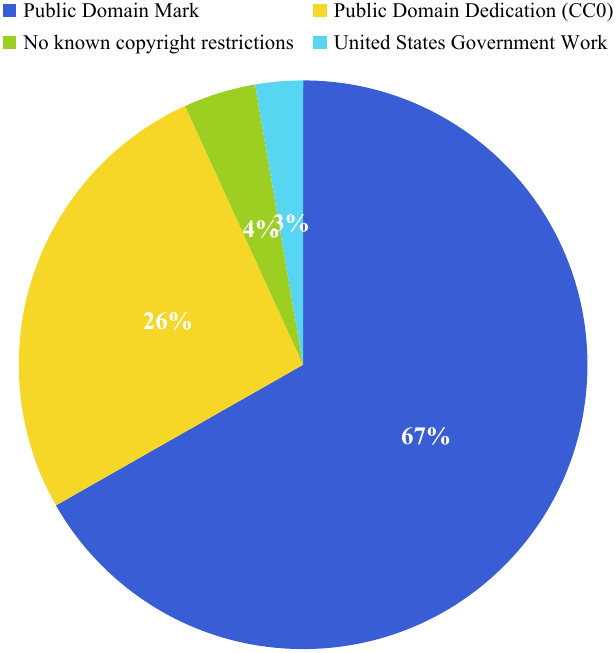}
    \caption{Distribution of copyright statuses for real images in OpenSDID.}
    \label{fig:copyright}
\end{figure}

\noindent\textbf{Copyright Considerations.}
To ensure the dataset is publicly accessible, we have made every effort to ensure that \zhiwu{all locally manipulated images based on ~\cite{BoerBohan2024Megalith10m}} are free from copyright restrictions.
All images used in this research are freely available under one or more of the following licenses: No Known Copyright Restrictions, United States Government Work, Public Domain Dedication (CC0), or Public Domain Mark. These licenses permit unrestricted use, modification, and distribution of the visual materials. Specifically, the images fall into the following categories: (a) materials free from known copyright restrictions due to their age or provenance, (b) content created by U.S. federal government employees as part of their official duties, (c) works explicitly dedicated to the public domain through Creative Commons Zero designation, or (d) materials marked as public domain due to copyright expiration or other legal provisions. Although these licenses do not require attribution, we have included source citations where applicable to maintain academic integrity and adhere to best practices.
The distribution of these copyright statuses is shown in Figure~\ref{fig:copyright}.

However, it is important to acknowledge that, given the nature of the Internet, there remains a possibility that a small number of images may have been uploaded by users without proper copyright clearance.
While we reserve all rights to the AI-generated images produced in this research, we do not claim ownership of the original source images. Researchers can access these source images independently through the image URLs we have provided in our dataset.

\noindent\textbf{SFW Ensurance.}
Our dataset consists of images retrieved through the Flickr API, implementing rigorous safety protocols with maximum safety settings (safety\_level=1) to ensure content appropriateness~\cite{BoerBohan2024Megalith10m}. 
For more comprehensive safety verification, particularly for artificially generated images, we employed multiple state-of-the-art NSFW detectors including:
\begin{itemize}
\item GantMAN NSFW Detector~\cite{laborde2019nsfw}
\item LAION's CLIP-based NSFW Detector~\cite{schuhmann2022laionsafety}
\item Stable Diffusion Safety Checker~\cite{rando2022redteaming}
\end{itemize}
Our multi-layered verification process confirmed that 100\% of the images maintain SFW status, which can be attributed to both the strict initial filtering and the inherent safety mechanisms in the underlying VLMs and T2I generators.

\noindent\textbf{Potential Biases Analysis.}
To conduct a comprehensive bias analysis of our dataset, we employed two sophisticated detection frameworks: EasyFace~\cite{sithu31296_easyface} for human demographic analysis and Florence-2 for general object detection.

We utilized the EasyFace detection framework to assess representation across multiple demographic dimensions. Our analysis encompassed binary gender classification, seven distinct racial/ethnic categories, and nine age groups spanning from infancy to elderly, providing a granular view of demographic representation within the dataset.
The face detection algorithm was applied across the entire dataset, utilizing pre-trained models optimized for multi-attribute recognition. 
Table~\ref{tab:demographics} presents the results. 
While this analysis provides valuable insights into dataset representation, we acknowledge the inherent limitations of automated demographic classification systems, particularly when dealing with edge cases and intersectional identities.

\begin{figure*}
    \centering
    \includegraphics[width=1\linewidth]{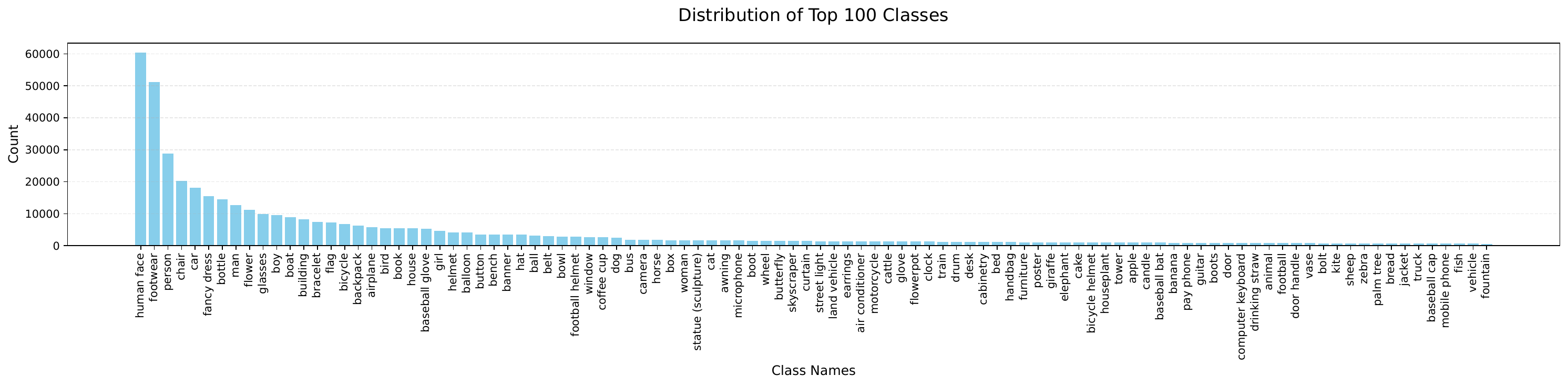}
    \caption{\zhiwu{Distribution of top-100 classes in the OpenSDID}}
    \label{fig:class_distribution}
\end{figure*}

To complement the demographic analysis, we employed Florence-2 for open-world object detection, providing insights into the distribution of general categories within the dataset. The results are visualized in Figure~\ref{fig:class_distribution}.

\begin{table}[t]
\centering
\begin{tabular}{llr}
\toprule
\textbf{Category} & \textbf{Attribute} & \textbf{Percentage (\%)} \\
\midrule
\multirow{2}{*}{Gender}
& Male & 53.4 \\
& Female & 46.6 \\
\midrule
\multirow{7}{*}{Race/Ethnicity}
& White & 46.5 \\
& Black & 7.0 \\
& Latino Hispanic & 4.4 \\
& East Asian & 20.5 \\
& Southeast Asian & 2.4 \\
& Indian & 1.7 \\
& Middle Eastern & 17.5 \\
\midrule
\multirow{9}{*}{Age}
& 0-2 & 0.4 \\
& 3-9 & 6.4 \\
& 10-19 & 6.7 \\
& 20-29 & 36.4 \\
& 30-39 & 18.6 \\
& 40-49 & 14.2 \\
& 50-59 & 11.6 \\
& 60-69 & 5.0 \\
& 70+ & 0.7 \\
\bottomrule
\end{tabular}
\caption{Demographic distribution analysis results \zhiwu{on the OpenSDID dataset}.}
\label{tab:demographics}
\end{table}

\section{More Related Work Discussion}
\label{sec:related_works_discussion}

For training and evaluation, we use the implementation of the IMDL-BenCo framework~\cite{ma2024imdl}, which includes several state-of-the-art methods for comparison. Here, we briefly describe the methods included in our paper.

CAT-Net (Compression Artifact Tracing Network)~\cite{kwon2022learning} is a dual-stream neural network that simultaneously processes both RGB images and JPEG compression artifacts (DCT coefficients) to detect image manipulations. By leveraging both visual content and compression artifacts, it is particularly effective at identifying forgeries, even in compressed images.

MVSS-Net (Multi-View Multi-Scale Supervised Networks)~\cite{chen2021image} is a dual-branch architecture that simultaneously processes both RGB domain features and noise patterns. The edge-supervised branch utilizes edge residual blocks, while the multi-scale feature learning branch analyzes tampering edge artifacts and noise views of input images.

PSCC-Net (Progressive Spatio-Channel Correlation Network)~\cite{liu2022pscc} leverages multi-scale feature learning through dense cross-connections and progressive feature fusion strategies. The model utilizes different sizes of convolutions and perceptual fields to extract valuable information about tampered locations, making it particularly effective at detecting various types of image manipulations.

TruFor~\cite{guillaro2023trufor} uses a Noiseprint++ extractor to process RGB images and obtain learned noise-sensitive fingerprints. These fingerprints, along with the original RGB input, are fed into an encoder that jointly computes features for two parallel decoding paths: an anomaly decoder for pixel-level forgery localization and a confidence decoder for detection.

ObjectFormer~\cite{wang2022objectformer} is a transformer-based architecture designed for image manipulation detection and localization. It takes both RGB domain and frequency domain (DCT) as input.

IML-ViT~\cite{ma2024imlvit} employs a specialized architecture that combines a windowed Vision Transformer (ViT) backbone, which alternates between windowed and global attention blocks to process high-resolution ($1024 \times 1024$) input images, with a Simple Feature Pyramid Network (SFPN) for multi-scale feature extraction.

DeCLIP~\cite{smeu2024declip} leverages CLIP ViT-L/14 as its image encoder and employs a convolutional-based architecture as its decoder. The CLIP image encoder is kept frozen, while only the mask decoder is trained for deepfake localization.

CNNDet~\cite{wang2020cnn} is a standard image classifier trained on images generated by a single CNN generator (ProGAN). Through careful data augmentations, it can successfully detect AI-generated images across multiple different architectures and datasets.

UniFD~\cite{ojha2023towards} detects AI-generated fake images by extracting features from a frozen CLIP-ViT model and then classifying these features using either nearest neighbor classification or an MLP classifier.

NPR~\cite{tan2023rethinking} detects synthetic images by analyzing patterns in the relationships between neighboring pixels, which result from upsampling operations in generative networks.

GramNet~\cite{liu2020global} detects fake faces by analyzing global texture patterns in images, utilizing a specialized neural network that extracts and compares statistical features from real and AI-generated facial textures.

FreqNet~\cite{tan2024frequencyaware} integrates a high-frequency representation module and frequency convolutional layer into a lightweight CNN architecture, processing both phase and amplitude spectra between FFT and IFFT operations while forcing continuous focus on high-frequency information to effectively detect deepfakes.

RINE~\cite{10.1007/978-3-031-73220-1_23} detects AI-generated images by extracting features from multiple intermediate layers of CLIP's Vision Transformer. These extracted features are then used to train a binary classifier that distinguishes between real and fake images.

Methods like CAT-Net and MVSS-Net significantly enhance detection accuracy through their sophisticated integration of multi-domain data, including RGB images, frequency domain information, and noise patterns. Approaches such as ObjectFormer and IML-ViT leverage transformer-based architectures to achieve precise detection and localization capabilities. More recent innovations, including DeCLIP and UniFD, harness the power of pre-trained CLIP models for robust feature extraction, pairing them with specialized decoders or classifiers to achieve state-of-the-art detection performance.

The open-world nature of the \taskname{} introduces several challenges. First, the diversity of user preferences and the constant innovation in diffusion models make it difficult to train models that generalize well. Second, the wide range of manipulation scopes, from global image synthesis to local edits, requires models to be robust across different scales and types of modifications.

\methodname{} advances beyond existing approaches through several key architectural innovations. Unlike methods such as CAT-Net and MVSS-Net that primarily focus on specific artifacts or dual-stream architectures, \methodname{} leverages a more comprehensive approach through its SPM framework. This framework uniquely combines CLIP's semantic understanding capabilities with MAE's reconstruction power, creating a more robust foundation for detection and localization tasks.

Though some previous studies also leverage pre-trained models, such as CLIP \cite{smeu2024declip, 10.1007/978-3-031-73247-8_2} and SAM \cite{lai2023detect}, we integrate both CLIP and MAE to achieve more generalized outcomes. 
The efficacy of our design is substantiated through comprehensive experimental validation on the OpenSDID dataset, illustrating notable enhancements over current methods in terms of both detection accuracy and localization precision.

\section{More Quantitative Results}
\label{sec:more_results}

In this section, \ZH{in addition to assessing the efficiency and the scalability of MaskCLIP and other state-of-the-art methods}, we focus on conducting a series of key additional experiments to examine the performance and capabilities of our proposed method, \methodname{}, thoroughly across various domains and scenarios. These experiments aim to provide a comprehensive understanding of \methodname{}'s effectiveness in diverse contexts, ranging from traditional image manipulation detection to more advanced challenges posed by AI-generated content.

\ZH{The main paper focuses on studying the proposed OpenSDI challenge and evaluates crossing the constructed OpenSDI datasets (e.g., SD1.5, Flux.1). In contrast, the evaluation across OpenSDI and non-OpenSDI (other public) datasets is used to study the broader benefits of our OpenSDI dataset and therefore is included in the suppl. material (Tables~\ref{tab:genimage} and \ref{table:zeroshot_eval}).}

\noindent \textbf{Scalability Analysis.}
\yabin{To evaluate the data scalability of MaskCLIP, we conducted experiments by training MaskCLIP and the SOTA method TruFor on different proportions of the OpenSDID dataset. Table \ref{tab:comparison_results1} presents the pixel-level F1 scores for both methods when trained on 25\%, 50\%, and 100\% of the data. 
As shown, the performance of MaskCLIP improves as the training data size increases from 25\% to 100\%, demonstrating its favorable data scaling properties and ability to leverage larger datasets for enhanced performance in OpenSDI tasks.}

\begin{table}[h]
\centering
\footnotesize
\begin{tabular}{c|cc}
\hline
Data Proportion & TruFor & MaskCLIP \\
\hline
25\% & 0.5294 & 0.5737 \\
50\% & 0.6062 & 0.5906 \\
100\% & 0.7100 & 0.7563 \\
\hline
\end{tabular}
\caption{Pixel-level F1 scores of MaskCLIP \& TruFor (SOTA) on different proportions of the OpenSDID dataset.}
\label{tab:comparison_results1}
\end{table}

\noindent\textbf{Cross-Dataset Evaluation on Traditional Image Forgery Detection and Localization (IMDL) Benchmarks.} Table~\ref{tab:cross_dataset_comparison} presents a comprehensive comparison between our proposed \methodname{} method and state-of-the-art image manipulation detection methods across five established forensics datasets (COVERAGE, Columbia, NIST16, CASIAv1, and IMD2020). 
Following the IMDL-BenCo framework~\cite{ma2024imdl}, we adopt Protocol-MVSS, where all models are trained exclusively on the CASIAv2 dataset and evaluated directly on other datasets without fine-tuning, enabling a true assessment of zero-shot domain generalization capabilities.
The experimental results demonstrate the superior performance of our method across most evaluation scenarios. Notably, our approach achieves the highest average F1 score and outperforms existing methods on three out of five benchmarks. It is important to note that all test benchmarks focus on traditional image manipulation detection tasks, specifically addressing conventional manipulation techniques such as copy-paste and splicing operations. The primary variations across these datasets stem from differences in image content domains and resolutions.
But our \taskname{} challenge tackles a fundamentally different problem: detecting and localizing manipulations generated by advanced AI models, particularly diffusion-based T2I generation methods. 

\begin{table*}
\centering
\begin{tabular}{l|cccccc}
\hline
Method & COVERAGE & Columbia & NIST16 & CASIAv1 & IMD2020 & Average \\
\hline
Mantra-Net~\cite{Wu_2019_CVPR} & 0.090 & 0.243 & 0.104 & 0.125 & 0.055 & 0.123 \\
MVSS-Net~\cite{chen2021image} & 0.259 & 0.386 & 0.246 & 0.534 & 0.279 & 0.341 \\
CAT-Net~\cite{kwon2022learning} & 0.296 & 0.584 & 0.269 & 0.581 & 0.273 & 0.401 \\
ObjectFormer~\cite{wang2022objectformer} & 0.294 & 0.336 & 0.173 & 0.429 & 0.173 & 0.281 \\
PSCC-Net~\cite{liu2022pscc} & 0.231 & 0.604 & 0.214 & 0.378 & 0.235 & 0.333 \\
NCL-IML~\cite{Zhou_2023_ICCV} & 0.225 & 0.446 & 0.260 & 0.502 & 0.237 & 0.334 \\
Trufor~\cite{guillaro2023trufor} & 0.419 & \textbf{0.865} & 0.324 & {0.721} & 0.322 & {0.530} \\
IML-ViT~\cite{ma2024imlvit} & {0.435} & 0.780 & 0.331 & {0.721} & 0.327 & 0.519 \\
\hline
MaskCLIP & \textbf{0.451} & {0.848} & \textbf{0.342} & \textbf{0.725} & {0.319} & \textbf{0.537} \\
\hline
\end{tabular}
\caption{Pixel-level F1 (localization) performance across traditional image forensics benchmark datasets. All methods are trained on CASIAv2 and do the cross-dataset evaluation on these benchmarks.}
\label{tab:cross_dataset_comparison}
\end{table*}

\noindent\textbf{Cross-Dataset Evaluation on Another AI-generated Image Benchmark Dataset (GenImage).}
To further evaluate the performance of our proposed method, we conduct experiments on the recently introduced GenImage benchmark. This benchmark allows us to compare methods trained on the GenImage SDv1.4 subset with those trained on our \datasetname{}. The evaluation is divided into two groups: in-dataset evaluation and cross-dataset zero-shot evaluation.

In the in-dataset evaluation, the methods are trained and tested on the same distribution. In contrast, the cross-dataset zero-shot evaluation testing models trained on \datasetname{} on GenImage dataset. 
Specifically, GenImage uses ImageNet as its real source data, while \datasetname{} is composed of web-collected images, introducing a significant domain gap between the two datasets.

The results in Table~\ref{tab:genimage} demonstrate not only our method's performance but also highlight \datasetname{}'s superiority in terms of image diversity and realism. The table presents accuracy metrics for various methods across different testing subsets, including Midjourney, SD V1.4, SD V1.5, ADM, GLIDE, Wukong, VQDM, and BigGAN, along with the average accuracy.
The results reveal that methods trained on \datasetname{} demonstrate superior performance in cross-dataset zero-shot evaluation, despite the domain gap. This suggests that \datasetname{} provides a more challenging and diverse training set, enhancing model generalization capabilities. Our proposed method, MaskCLIP, achieves a competitive average accuracy of 77.4\%, surpassing several state-of-the-art methods in the cross-dataset setting.

\begin{table*}
\begin{tabular}{l|cccccccc|c}
\hline
\multirow{2}{*}{Method} & \multicolumn{8}{c|}{Testing Subset} & \multirow{2}{*}{Avg} \\
\cline{2-9}
& Midjourney & SD V1.4 & SD V1.5 & ADM & GLIDE & Wukong & VQDM & BigGAN & Acc.(\%) \\
\hline
ResNet-50$^{\dagger}$ & 54.9 & \textbf{99.9} & 99.7 & 53.5 & 61.9 & 98.2 & 56.6 & 52.0 & 72.1 \\
DeiT-S$^{\dagger}$ & 55.6 & \textbf{99.9} & 99.8 & 49.8 & 58.1 & 98.9 & 56.9 & 53.5 & 71.6 \\
Swin-T$^{\dagger}$ & 62.1 & \textbf{99.9} & 99.8 & 49.8 & 67.6 & 99.1 & 62.3 & 57.6 & 74.8 \\
CNNDet$^{\dagger}$ & 52.8 & 96.3 & 95.9 & 50.1 & 39.8 & 78.6 & 53.4 & 46.8 & 64.2 \\
Spec$^{\dagger}$ & 52.0 & 99.4 & 99.2 & 49.7 & 49.8 & 94.8 & 55.6 & 49.8 & 68.8 \\
F3Net$^{\dagger}$ & 50.1 & \textbf{99.9} & \textbf{99.9} & 49.9 & 50.0 & \textbf{99.9} & 49.9 & 49.9 & 68.7 \\
GramNet$^{\dagger}$ & 54.2 & 99.2 & 99.1 & 50.3 & 54.6 & 98.9 & 50.8 & 51.7 & 69.9 \\
\midrule    
PSCC-Net~\cite{liu2022pscc} & 65.0 & 94.9 & 95.3 & 55.0 & 56.2 & 79.5 & 53.6 & 59.4 & 69.9 \\
MVSS-Net~\cite{chen2021image} & \textbf{66.7} & 84.4 & 85.2 & 60.3 & 62.9 & 69.8 & 67.8 & 70.0 & 70.9 \\
TruFor~\cite{guillaro2023trufor} & 55.6 & 46.0 & 45.0 & 66.0 & 60.8 & 51.0 & 70.7 & 70.4 & 58.2 \\
DeCLIP~\cite{smeu2024declip} & 56.3 & 79.8 & 79.0 & \textbf{72.4} & \textbf{75.5} & 79.9 & 77.5 & \textbf{81.3} & 75.2 \\
\hline
MaskCLIP & 51.5 & 95.0 & 96.3 & 70.6 & \textbf{75.5} & 73.8 & \textbf{78.3} & 77.8 & \textbf{77.4} \\
\hline
\end{tabular}
\caption{\zhiwu{Image-level (detection accuracy) performance} on the GenImage benchmark dataset~\cite{zhu2024genimage}. The methods are divided into two groups based on their training data: the first group (marked with †) is trained on GenImage SDv1.4, while the second group (no marks) is trained on \datasetname{}. }
\label{tab:genimage}
\end{table*}

\noindent\textbf{Zero-shot Evaluation on \datasetname{} with Different Pretrained Dataset.}
We conduct comprehensive zero-shot evaluations to assess the generalization capability of various state-of-the-art forgery detection methods across different training settings. Table~\ref{table:zeroshot_eval} presents the pixel-level localization performance using IoU and F1 metrics on \datasetname{} test sets. We evaluate three representative methods: TruFor~\cite{guillaro2023trufor}, DeCLIP~\cite{smeu2024declip}, and IML-ViT~\cite{ma2024imlvit}, each tested with their officially released pretrained weights and trained on our \datasetname{}.

Several key observations emerge:
First, when using original pretrained weights, all methods show limited generalization to T2I diffusion-generated images, with performance declining significantly compared to their reported results on traditional forgery datasets. This indicates that existing methods trained on traditional manipulation data struggle to transfer to diffusion-based forgeries.
Second, retraining these methods on \datasetname{} substantially improves their performance, particularly for SD1.5 and SD2.1 models. For instance, TruFor's IoU improves from 0.0742 to 0.6342 on SD1.5 after retraining.
These results highlight both the challenge and importance of developing robust detection methods specifically designed for diffusion-based forgeries, as traditional datasets show limited effectiveness in this emerging threat landscape.

\begin{table*}
\centering
\resizebox{1\linewidth}{!}{
\begin{tabular}{l|l|cc|cc|cc|cc|cc|cc} \hline
\multirow{2}{*}{} & 
\multirow{1}{*}{} & \multicolumn{2}{c|}{SD1.5} & \multicolumn{2}{c|}{SD2.1} & \multicolumn{2}{c|}{SDXL} & \multicolumn{2}{c|}{SD3} & \multicolumn{2}{c|}{Flux.1} &
\multicolumn{2}{c}{AVG} \\
Method & Data & IoU & F1 & IoU & F1 & IoU & F1 &  IoU & F1 &  IoU & F1 & IoU & F1 \\
\hline
TruFor~\cite{guillaro2023trufor} & Trufor$^{\dagger}$ & 0.0742 & 0.1073 & 0.0770 & 0.1115 & 0.0704 & 0.1035 & 0.0996 & 0.1424 & 0.1019 & 0.1464 & 0.0846 & 0.1222 \\
TruFor~\cite{guillaro2023trufor} & \datasetname{} & 0.6342 & 0.7100 & 0.5467 & 0.6188 & 0.2655 & 0.3185 & 0.3229 & 0.3852 & 0.0760 & 0.0970 & 0.3691 & 0.4259 \\
\hline
DeCLIP~\cite{smeu2024declip} & Dolos-LDM & 0.0138 & 0.0218 & 0.0131 & 0.0210 & 0.0089 & 0.0144 & 0.0145 & 0.0230 & 0.0070 & 0.0115 & 0.0115 & 0.0183 \\
DeCLIP~\cite{smeu2024declip}  & Dolos-Lama & 0.0093 & 0.0151 & 0.0098 & 0.0158 & 0.0057 & 0.0098 & 0.0155 & 0.0251 & 0.0048 & 0.0085 & 0.0090 & 0.0149 \\
DeCLIP~\cite{smeu2024declip}  & Dolos-Pluralistic & 0.0145 & 0.0233 & 0.0154 & 0.0245 & 0.0064 & 0.0108 & 0.0182 & 0.0292 & 0.0069 & 0.0116 & 0.0123 & 0.0199 \\
DeCLIP~\cite{smeu2024declip} & Dolos-Repaint & 0.0254 & 0.0377 & 0.0221 & 0.0342 & 0.0184 & 0.0284 & 0.0344 & 0.0522 & 0.0162 & 0.0253 & 0.0233 & 0.0356 \\
DeCLIP~\cite{smeu2024declip} & \datasetname{} & 0.3718 & 0.4344 & 0.3569 & 0.4187 & 0.1459 & 0.1822 & 0.2734 & 0.3344 & 0.1121 & 0.1429 & 0.2520 & 0.3025 \\
\hline
IML-ViT~\cite{ma2024imlvit} & Trufor$^{\dagger}$ & 0.0806 & 0.1143 & 0.0825 & 0.1165 & 0.0746 & 0.1066 & 0.1279 & 0.1750 & 0.1295 & 0.1768 & 0.0990 & 0.1378 \\
IML-ViT~\cite{ma2024imlvit} & CASIAv2 & 0.0248 & 0.0384 & 0.0228 & 0.0366 & 0.0213 & 0.0337 & 0.0266 & 0.0418 & 0.0290 & 0.0460 & 0.0249 & 0.0393 \\
IML-ViT~\cite{ma2024imlvit} & \datasetname{} & 0.6651 & 0.7362 & 0.4479 & 0.5063 & 0.2149 & 0.2597 & 0.2363 & 0.2835 & 0.0611 & 0.0791 & 0.3251 & 0.3730 \\
\hline
\methodname & CASIAv2 & 0.0312 & 0.0465 & 0.0289 & 0.0442 & 0.0256 & 0.0398 & 0.0334 & 0.0502 & 0.0358 & 0.0532 & 0.0310 & 0.0468 \\
\methodname & \datasetname{} & 0.6712 & 0.7563 & 0.5550 & 0.6289 & 0.3098 & 0.3700 & 0.4375 & 0.5121 & 0.1622 & 0.2034 & 0.4271 & 0.4941 \\
\hline
\end{tabular}
}
\vspace{-0.2cm}
\caption{Pixel-level (localization) performance on \datasetname{}. Test SOTA methods pretrained weights on various training data and do zero-shot testing on the OpenSDID's test sets.
Trufor$^{\dagger}$ indicates training on a combined dataset including CASIA v2, FantasticReality~\cite{kniaz2019point}, IMD2020, and tampered versions of COCO and RAISE datasets~\cite{kwon2022learning}, which is the training setting of Trufor.
}
\label{table:zeroshot_eval}
\vspace{-0.1cm}
\end{table*}

\begin{figure}
    \centering
    \includegraphics[width=1\linewidth]{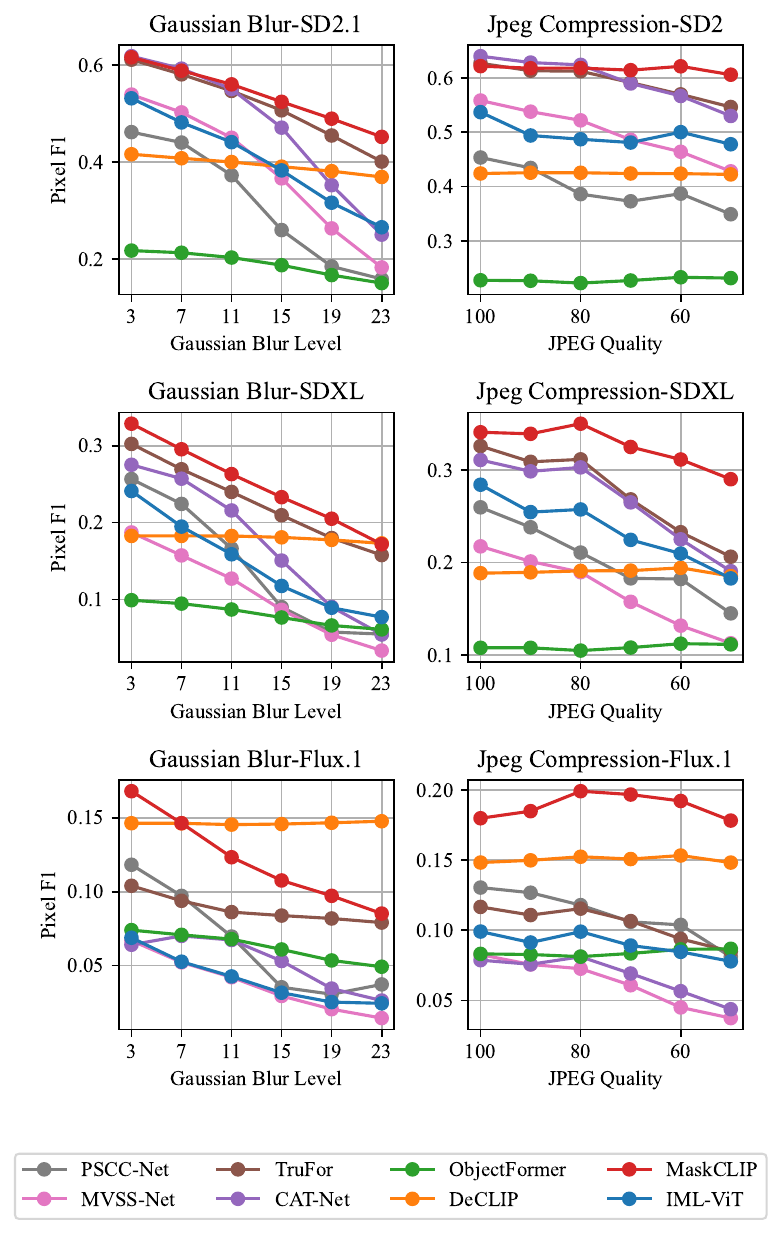}
    \caption{Robustness evaluation of different SOTA methods under image degradation on OpenSDID. It compares performance across varying levels of Gaussian Blur (left) and JPEG Compression (right).}
    \label{fig:more_robust}
\end{figure}

\noindent\textbf{More Robustness Analysis.}
Figure~\ref{fig:more_robust} shows the performance comparison of different methods under image degradation conditions, specifically Gaussian blur and JPEG compression, across three datasets (SD2.1, SDXL, and Flux.1). 
Under varying levels of Gaussian blur (3-23) and JPEG compression quality (60-100), MaskCLIP consistently demonstrates superior robustness compared to other approaches.
Particularly notable is MaskCLIP's performance on the SD2.1 dataset, where it maintains strong F1 scores even as image quality degrades. While performance naturally decreases with more severe degradation, MaskCLIP exhibits more graceful degradation compared to competing methods, maintaining its lead across different datasets and degradation types. This consistent performance advantage highlights the robust nature of our approach in handling various real-world image quality challenges.

\noindent\textbf{Comprehensive Analysis of Pretrained Model Combinations.}
Since our proposed Synergizing Pretrained Models (SPM) learning scheme leverages multiple pretrained models, with MaskCLIP serving as just one implementation example, we conducted extensive experiments with various model combinations. We explored different CLIP variants (ViT-B/32, OpenCLIP) and alternative pixel-wise encoders to thoroughly evaluate the SPM approach.
Table~\ref{tab:encoder_comparison} presents a systematic comparison of different encoder combinations. The OpenAI ViT-B/32 paired with MAE-base achieves F1 scores of 0.7227 and 0.4472 on SD1.5 and SD3 datasets, respectively. Scaling up to OpenCLIP ViT-L/14 yields improved performance, demonstrating the advantages of a larger vision transformer architecture.
Notably, substituting MAE-base with Dinov2-base results in decreased performance (F1 scores of 0.6278 and 0.3521), suggesting that MAE's self-supervised pretraining approach is more effective for our specific task. The optimal performance is achieved by combining OpenAI ViT-L/14 with SAM-base, reaching F1 scores of 0.7873 and 0.5773. This superior performance indicates that SAM's segment-focused pretraining provides particularly valuable features for our segmentation task, albeit at the cost of increased computational overhead during both training and inference.
These comprehensive experiments underscore two key findings: (1) the critical importance of selecting appropriate pretrained weights, and (2) the significant performance benefits that can be achieved through larger model architectures and segment-aware pretraining strategies.

\begin{table}
    \centering
    \begin{tabular}{llccc}
        \hline
        \multicolumn{2}{c}{Method} & & \multicolumn{2}{c}{Pixel-level F1 } \\
        \cline{1-2} \cline{4-5}
        Encoder1 & Encoder2 & Params. & SD1.5 & SD3 \\
        \hline
        OA ViT-B/32 & MAE-base & 96M & 0.7227 & 0.4472 \\
        OC ViT-L/14 & MAE-base & 114M  & 0.7363 & 0.4908 \\
        OA ViT-L/14 & Dinov2-base & 114M  & 0.6278 & 0.3521 \\
        OA ViT-L/14 & MAE-base & 114M  & 0.7563 & 0.3700 \\
        OA ViT-L/14 & SAM-base & 126M & \textbf{0.7873} & \textbf{0.5773} \\
        \hline
    \end{tabular}
    \caption{Performance comparison of different encoder combinations. We evaluate various pretrained models as Encoder1 (CLIP variants) and Encoder2 (pixel-wise encode models). OA and OC denote OpenAI and OpenCLIP respectively. The results show pixel-level F1 scores on both SD1.5 and SD3 datasets. Params. indicates the trainable parameters of the model, where we only keep the CLIP model frozen. }
    \label{tab:encoder_comparison}
\end{table}

\begin{table}
\centering
\begin{tabular}{ccc|cc}
\hline
\multicolumn{3}{c|}{Loss Components} & \multicolumn{2}{c}{Pixel-level F1} \\
\hline
$\mathcal{L}_{\text{CE}}$ & $\mathcal{L}_{\text{BCE}}$ & $\mathcal{L}_{\text{EDG}}$ & SD1.5 & SD3 \\
\hline
1.0 & 2.0 & 1.0 & 0.7575 & 0.3573 \\
2.0 & 1.0 & 1.0 & 0.7620 & 0.3373 \\
1.0 & 1.0 & 0.0 & 0.7299 & 0.3108 \\
1.0 & 1.0 & 1.0 & 0.7563 & 0.3700 \\
\hline
\end{tabular}
\caption{Ablation study on three loss function components. $\mathcal{L}_{\text{CE}}$ denotes cross-entropy loss, $\mathcal{L}_{\text{BCE}}$ represents binary cross-entropy loss, and $\mathcal{L}_{\text{EDG}}$ is the edge-weighted loss.}
\label{tab:loss_ablation}
\end{table}

\noindent\textbf{Analysis of Loss Function Components.}
The loss function in \methodname{} follows established approaches from previous works~\cite{ma2024imlvit, chen2021image, guillaro2023trufor}, incorporating three components: cross-entropy loss ($\mathcal{L}_{\text{CE}}$), binary cross-entropy loss ($\mathcal{L}_{\text{BCE}}$), and edge-weighted loss ($\mathcal{L}_{\text{EDG}}$). While we employ a simple balanced weighting scheme for these loss terms, it is crucial to analyze the impact of each component in our objective function. Table~\ref{tab:loss_ablation} presents experimental results with various weight combinations of the three loss terms. Our findings underscore the importance of maintaining a balanced objective function that appropriately weights binary classification and edge preservation.

\noindent\textbf{Multiple Runs Analysis.} Although most previous works don't perform multiple runs to study robustness, we conducted three independent training runs for both \methodname{} and the baseline TruFor method. As shown in Table~\ref{table:multipleruns}, \methodname{} exhibits notably smaller standard deviations across all subsets, indicating more stable and reliable performance compared to the baseline.

\begin{table}
\centering
\resizebox{1\linewidth}{!}{
\begin{tabular}{l|c|c|c|c|c|c} \hline
Method & SD1.5 & SD2.1 & SDXL & SD3 & Flux.1 & AVG \\
\hline
TruFor~\cite{guillaro2023trufor} & $0.728_{\pm 0.032}$ & $0.601_{\pm 0.029}$ & $0.289_{\pm 0.059}$ & $0.330_{\pm 0.018}$ & $0.071_{\pm 0.018}$ & $0.404_{\pm 0.031}$ \\
\hline
\methodname & $0.765_{\pm 0.008}$ & $0.628_{\pm 0.021}$ & $0.381_{\pm 0.013}$ & $0.492_{\pm 0.019}$ & $0.188_{\pm 0.009}$ & $0.471_{\pm 0.014}$ \\
\hline
\end{tabular}
}
\vspace{-0.2cm}
\caption{Performance of TruFor and \methodname{} with multiple runs on OpenSDID. }
\label{table:multipleruns}
\vspace{-0.1cm}
\end{table}

\section{More Qualitative Results}
\label{sec:sample1}

\begin{figure*}
    \centering
    \includegraphics[width=1\linewidth]{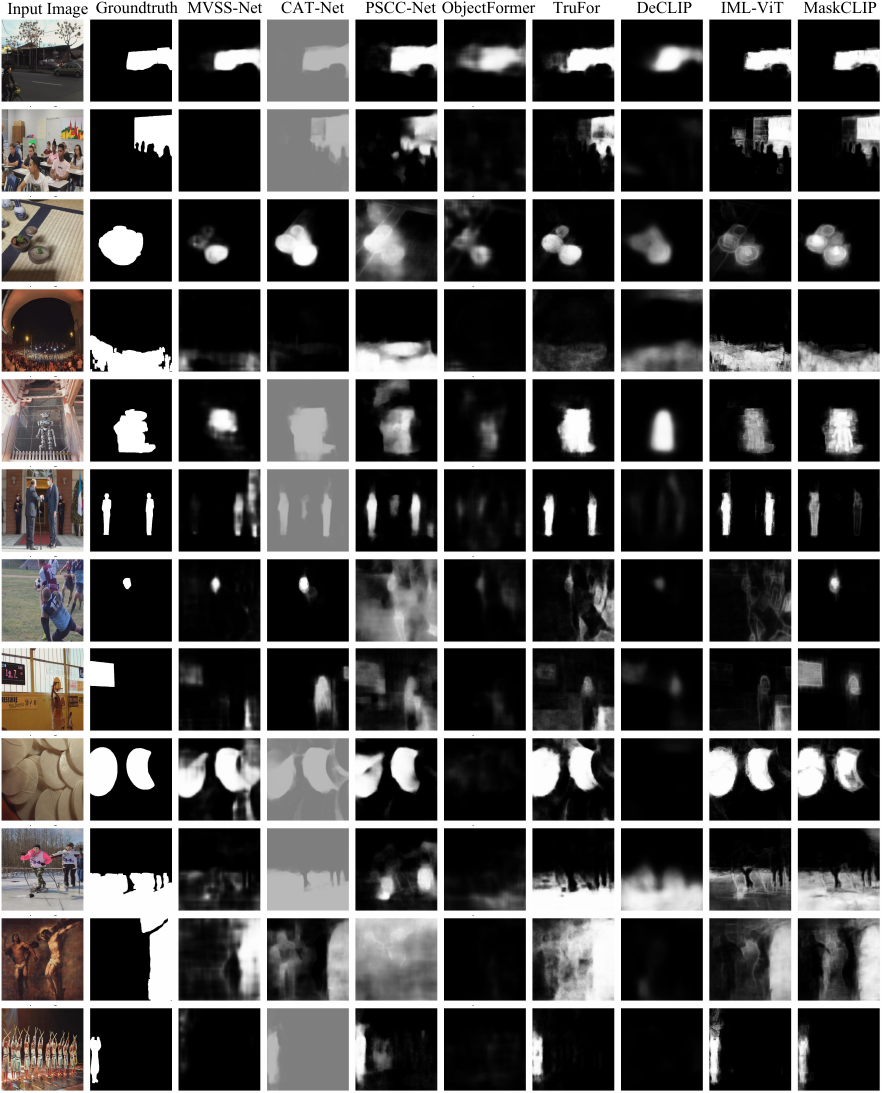}
    \caption{Qualitative results on \zhiwu{the OpenSDID SD1.5 testset}.}
    \label{fig:c-sd15}
\end{figure*}

\begin{figure*}
    \centering
    \includegraphics[width=1\linewidth]{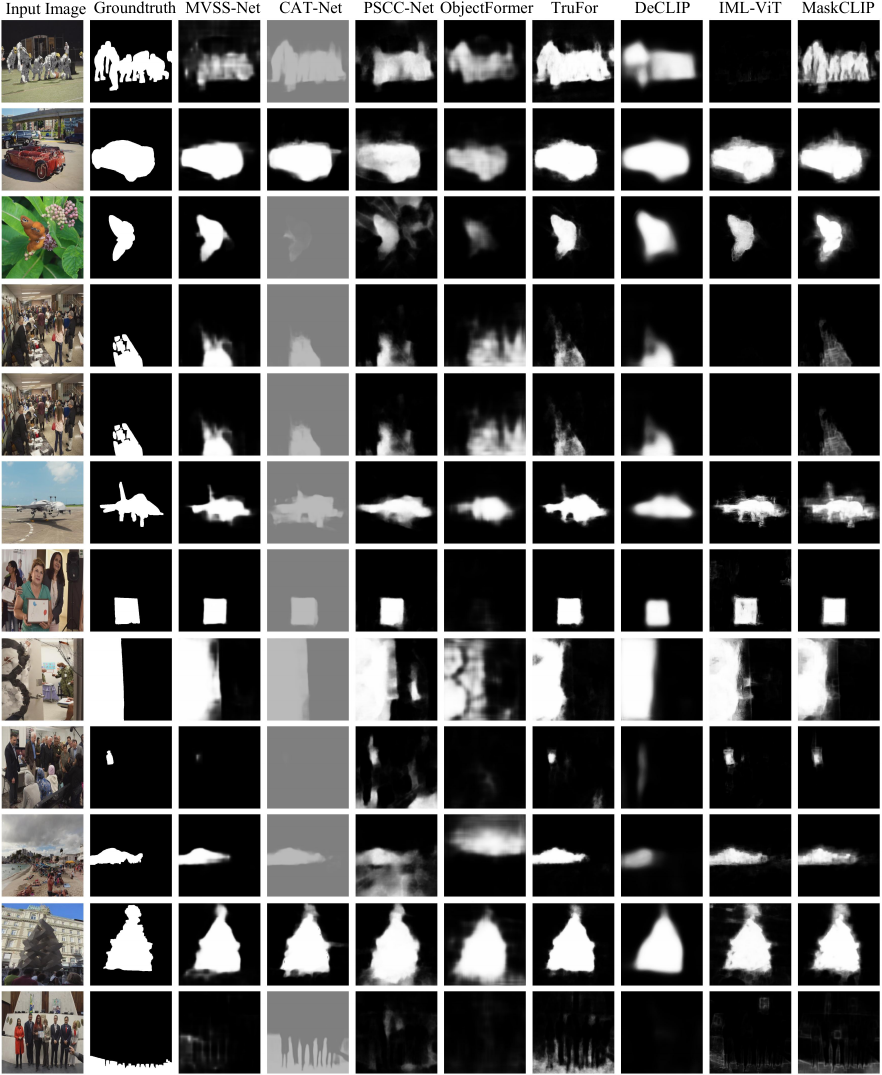}
    \caption{Qualitative results on \zhiwu{the OpenSDID SD2.1 test set}.}
    \label{fig:c-sd2}
\end{figure*}

\begin{figure*}
    \centering
    \includegraphics[width=1\linewidth]{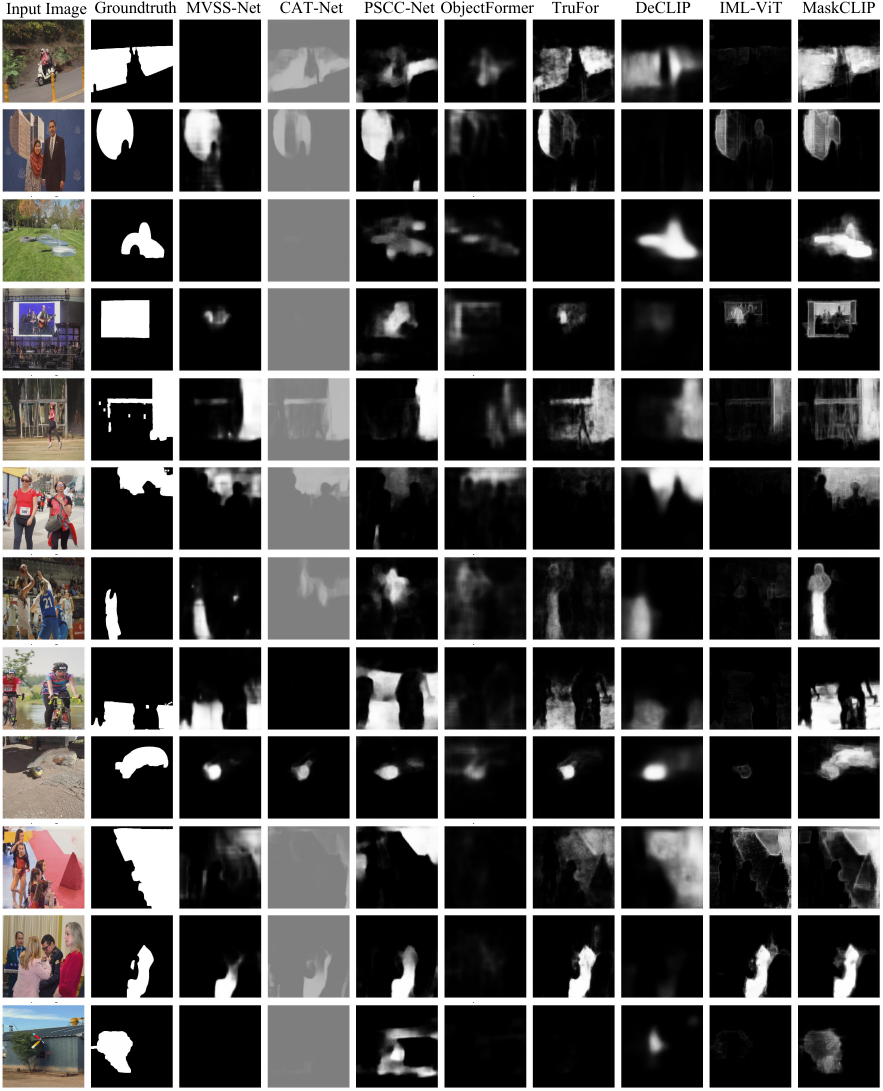}
    \caption{Qualitative results on \zhiwu{the OpenSDID SD3 test set}.}
    \label{fig:c-sd3}
\end{figure*}

\begin{figure*}
    \centering
    \includegraphics[width=1\linewidth]{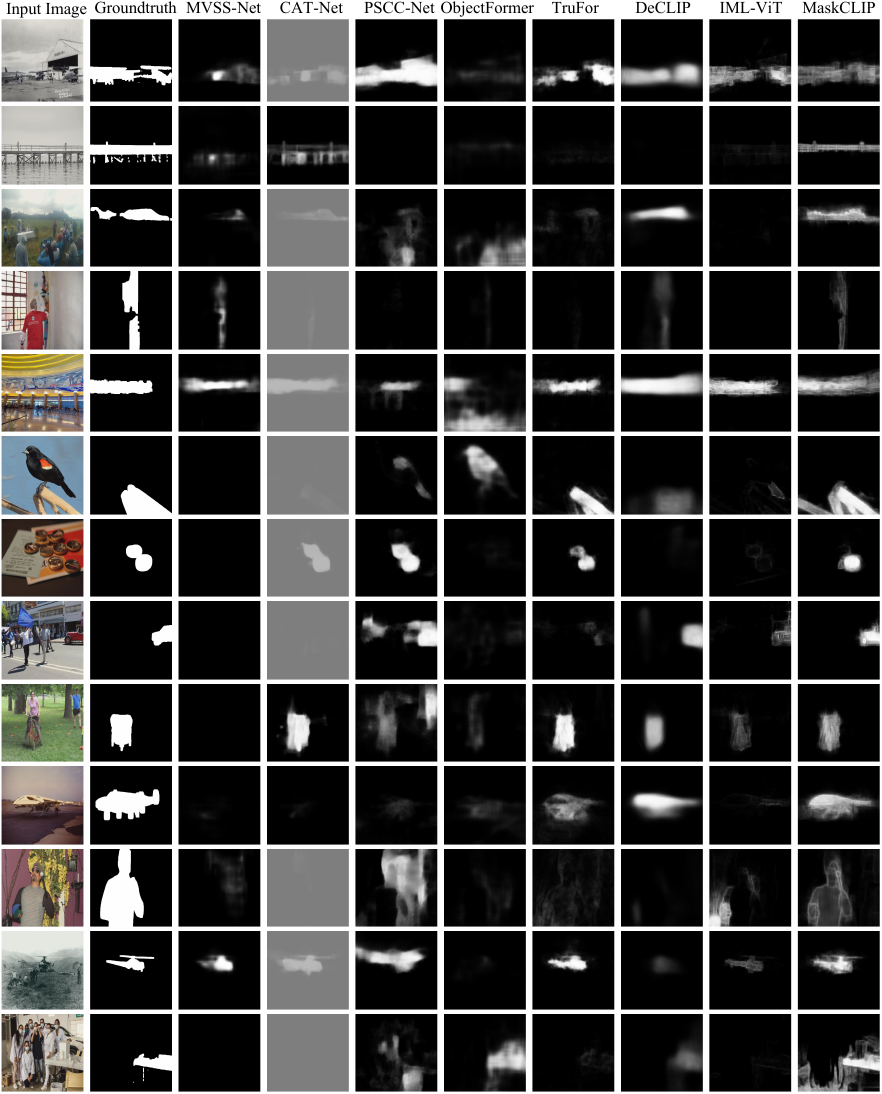}
    \caption{Qualitative results on \zhiwu{the OpenSDID SDXL test set}.}
    \label{fig:c-sdxl}
\end{figure*}

\begin{figure*}
    \centering
    \includegraphics[width=1\linewidth]{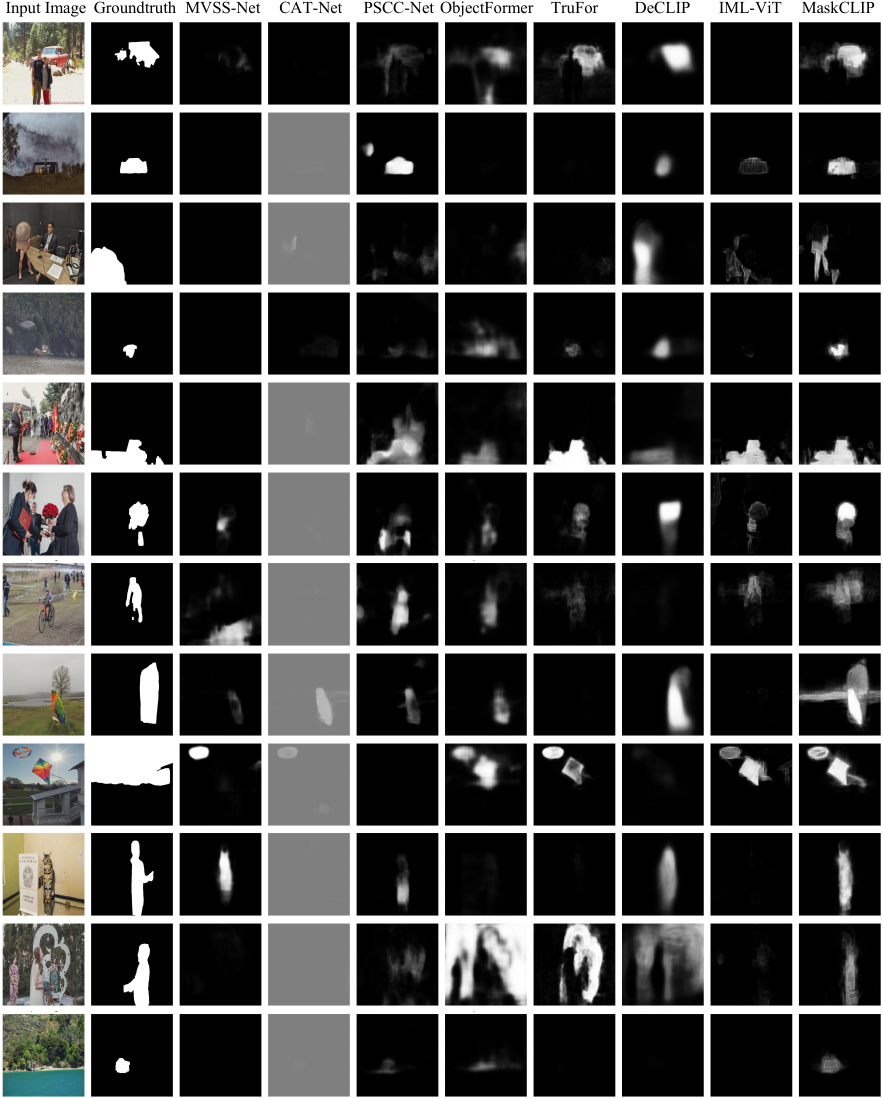}
    \caption{Qualitative results on \zhiwu{the OpenSDID Flux.1 test set}.}
    \label{fig:c-flux}
\end{figure*}

To provide a comprehensive evaluation of our method across different Diffusion models, we present additional qualitative results on various subsets of the OpenSDID dataset. Figures \ref{fig:c-sd15}, \ref{fig:c-sd2}, \ref{fig:c-sd3}, \ref{fig:c-sdxl}, and \ref{fig:c-flux} demonstrate our approach's performance on images generated by SD1.5, SD2.1, SD3, SDXL, and Flux.1 models respectively.

\begin{figure*}
    \centering
    \includegraphics[width=1\linewidth]{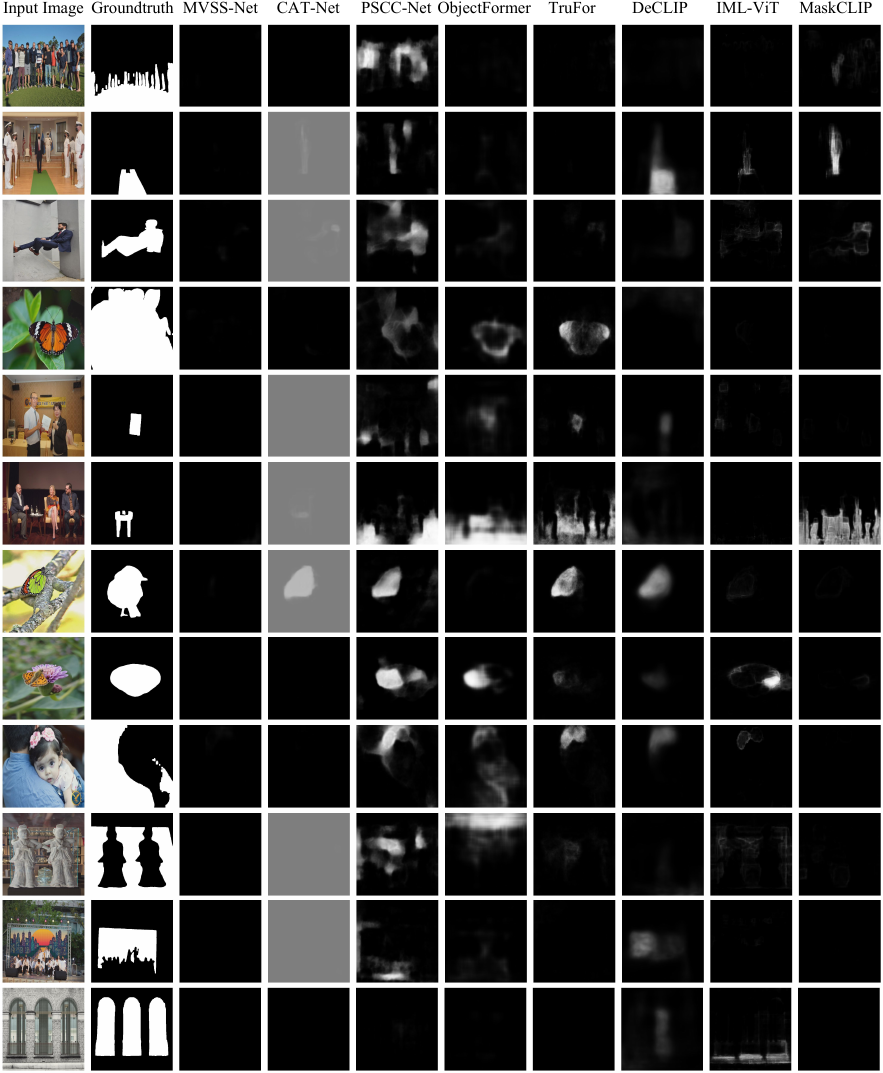}
    \caption{Some Fail Cases on \zhiwu{the OpenSDID Flux.1 test set}.}
    \label{fig:c-flux2}
\end{figure*}

The results showcase our method's effectiveness across various generator models. For the Flux.1 subset (Figure \ref{fig:c-flux}), our method exhibits strong zero-shot performance. However, in the interest of transparency, we also present several failure cases (Figure \ref{fig:c-flux2}) that highlight our method's current limitations. These challenging cases typically involve highly photorealistic images or compositions with unique elements that closely resemble natural photographs.
\yabin{These observations highlight the significant advancement of current AI image generation technology. This rapid development poses higher demands on image discrimination techniques and indicates that future research needs to develop more robust detection methods.}

\section{Samples of \datasetname{}}
\label{sec:sample2}

\begin{figure*}
    \centering
\includegraphics[width=0.7\linewidth]{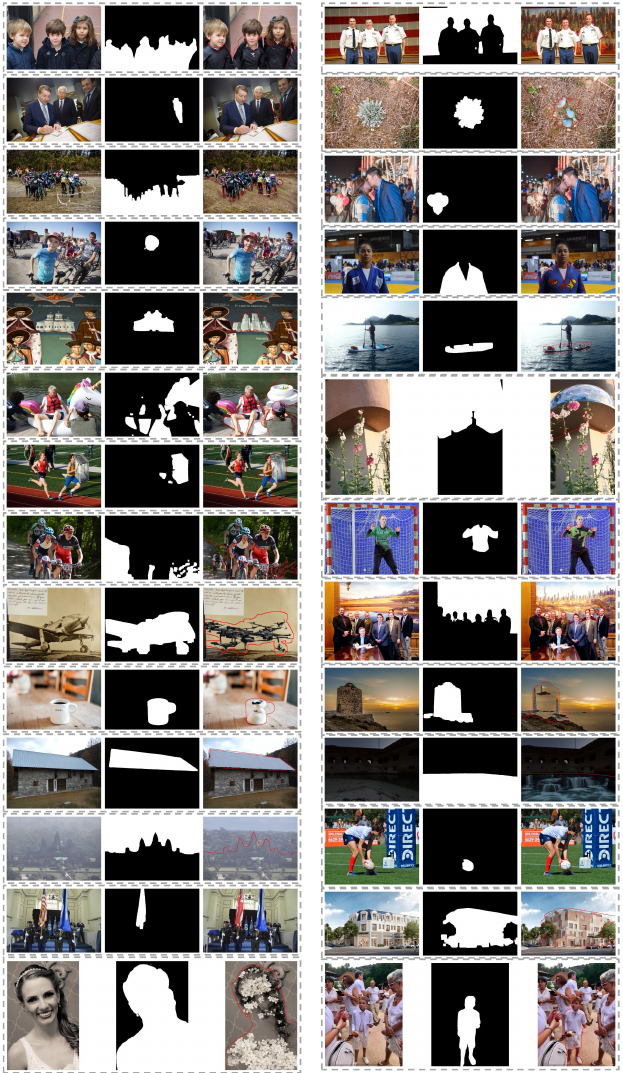}
    \caption{Sample images generated using the Stable Diffusion v1.5 (SD1.5) model.}
    \label{fig:sd15_samples}
\end{figure*}

\begin{figure*}
    \centering
\includegraphics[width=0.7\linewidth]{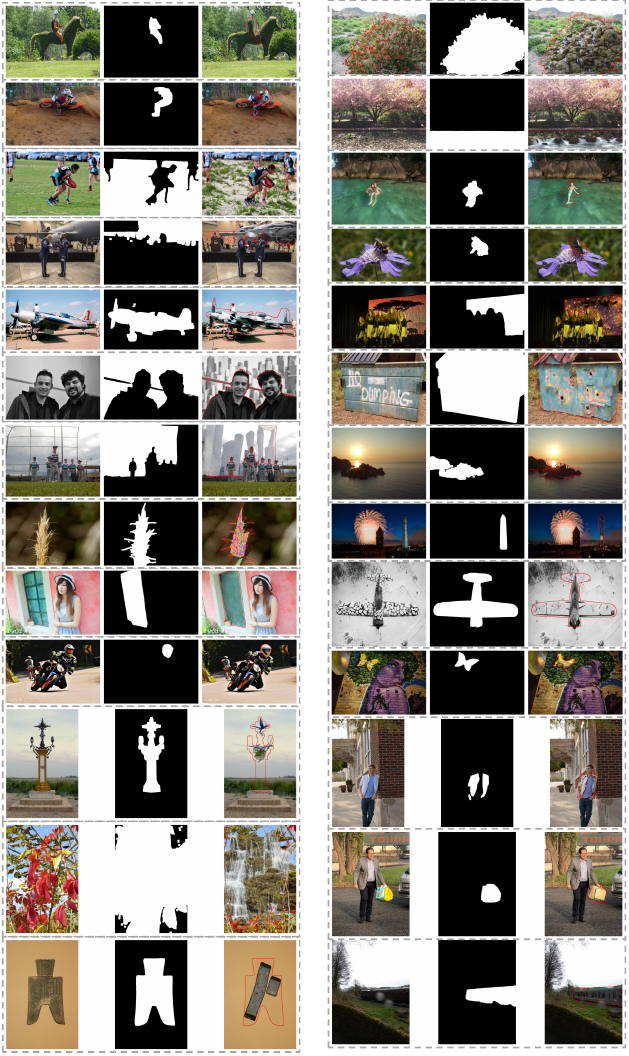}
    \caption{Sample images generated using the Stable Diffusion v2.1 (SD2.1) model.}
    \label{fig:sd2_samples}
\end{figure*}

\begin{figure*}
    \centering
\includegraphics[width=0.71\linewidth]{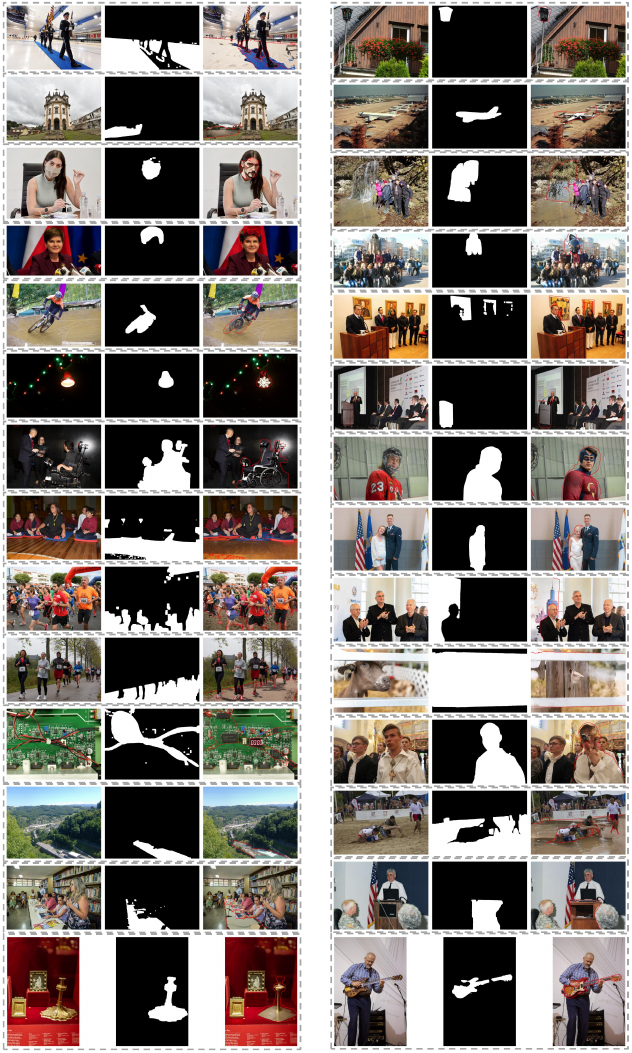}
    \caption{Sample images generated using the Stable Diffusion v3 model.}
    \label{fig:sd3_samples}
\end{figure*}

\begin{figure*}
    \centering
\includegraphics[width=0.7\linewidth]{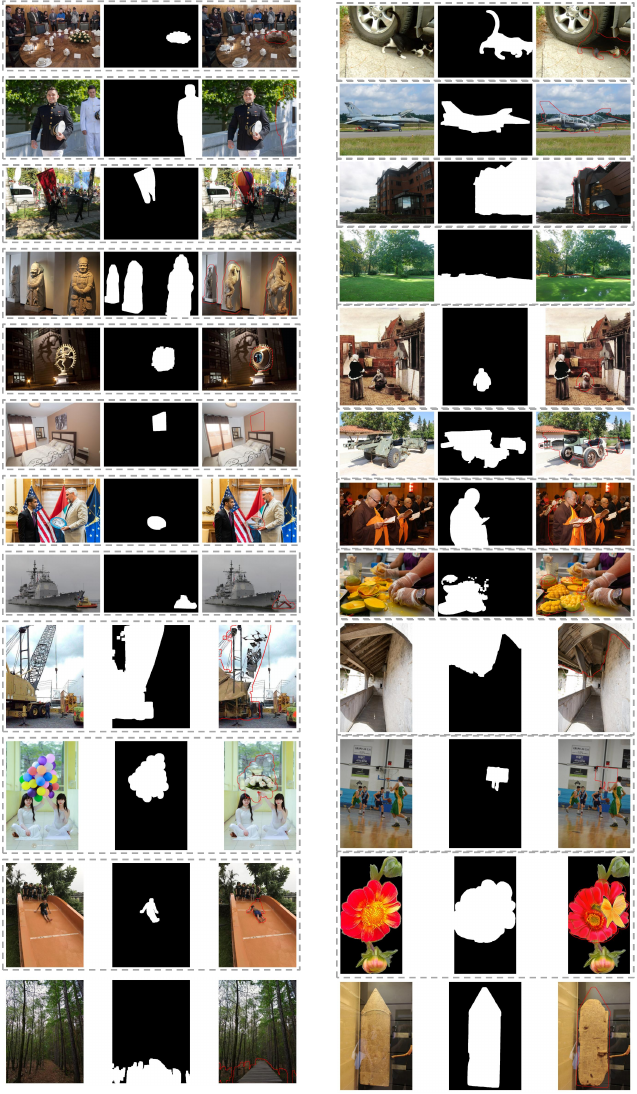}
    \caption{Sample images generated using the Stable Diffusion XL (SDXL) model.}
    \label{fig:sdxl_samples}
\end{figure*}

\begin{figure*}
    \centering
\includegraphics[width=0.7\linewidth]{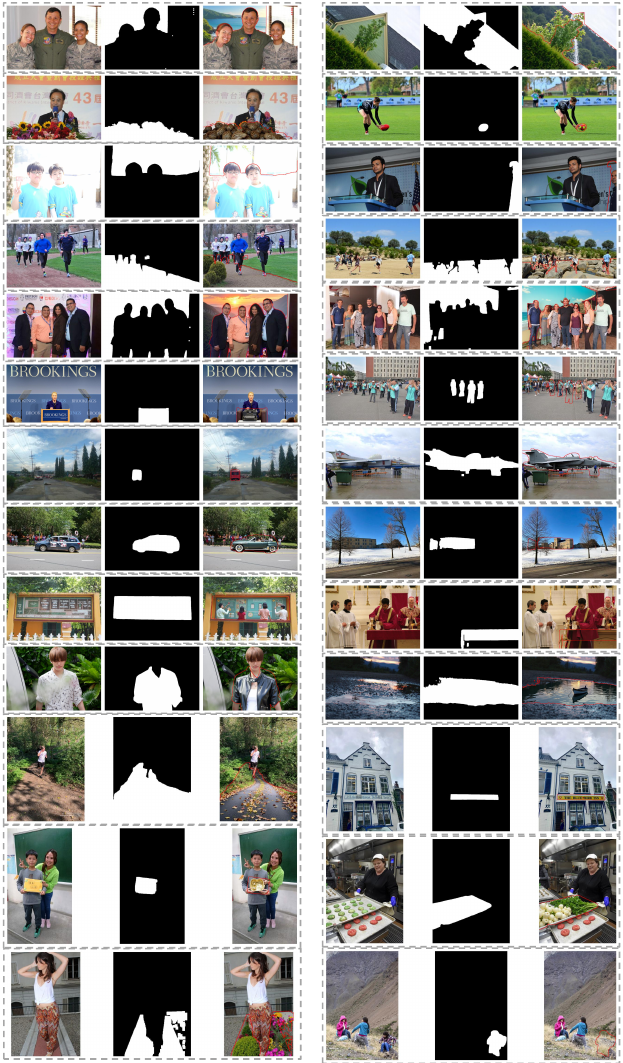}
    \caption{Sample images generated using the Flux.1 model.}
    \label{fig:flux_samples}
\end{figure*}

Figures~\ref{fig:sd15_samples}, Figures~\ref{fig:sd2_samples}, Figures~\ref{fig:sd3_samples}, Figures~\ref{fig:sdxl_samples}, and Figures~\ref{fig:flux_samples} present more samples of our datasets. 

{
    \small
    \bibliographystyle{ieeenat_fullname}
    \bibliography{main}
}